\documentclass[11pt]{article}

\usepackage[final]{acl}

\usepackage{times}
\usepackage{latexsym}
\usepackage{booktabs}
\usepackage[T1]{fontenc}
\usepackage[utf8]{inputenc}

\usepackage{microtype}

\usepackage{inconsolata}

\usepackage{graphicx}
\graphicspath{{../image/}{image/}}

\usepackage{amsmath}
\usepackage{tabularx}
\usepackage{makecell}
\usepackage{multirow}
\usepackage{booktabs}
\usepackage{enumitem}
\newcommand{\algo}{cross-modal skill injection}
\newcommand{\algohead}{Cross-modal skill injection}
\newcommand{\algoup}{Cross-Modal Skill Injection}

\title{Investigating Cross-Modal Skill Injection: Scenarios, Methods, and Hyperparameters}

\author{
  \textbf{Zhiyu Xu}\textsuperscript{1},
  \textbf{Lean Wang}\textsuperscript{1},
  \textbf{Yuanxin Liu}\textsuperscript{1},
  \textbf{Lei Li}\textsuperscript{3}, 
  \textbf{Hao Zhou}\textsuperscript{2},\\
  \textbf{Fandong Meng} \textsuperscript{2},
  \textbf{Jie Zhou}\textsuperscript{2},
  \textbf{Xu Sun}\textsuperscript{1} \\
  {\normalfont \textsuperscript{1}State Key Laboratory for Multimedia Information Processing,} \\
  {\normalfont School of Computer Science, Peking University} \\
  {\normalfont \textsuperscript{2}WeChat AI, Tencent Inc., China} 
  {\normalfont \textsuperscript{3}The University of Hong Kong} \\
  {\normalfont \texttt{zhiyu\_xu@stu.pku.edu.cn, lean@pku.edu.cn, liuyuanxin@stu.pku.edu.cn,}} \\
  {\normalfont \texttt{nlp.lilei@gmail.com, \{tuxzhou, fandongmeng, withtomzhou\}@tencent.com}} \\
  {\normalfont \texttt{xusun@pku.edu.cn}}
}

\usepackage{xcolor}
\usepackage{graphicx}

\begin{document}
\maketitle
\begin{abstract}

Vision-Language Models (VLMs) have demonstrated remarkable proficiency in general multi-modal understanding; yet they struggle to efficiently acquire continually evolving domain-specific skills. Conventional approaches to enhancing VLM capabilities, such as Supervised Fine-Tuning (SFT), require extensive dataset curation and substantial computational resources. Model merging has emerged as an efficient alternative that enables the transfer of domain-specific expertise from Large Language Models (LLMs) to VLMs without incurring additional training data requirements or significant computational overhead. Unlike conventional merging of homogeneous LLMs, which mainly aggregates existing capabilities, cross-modal skill injection aims to induce emergent cross-modal capabilities by integrating a domain-expert LLM into a VLM. However, existing research lacks a systematic analysis of the applicability and methodology of cross-modal skill injection. In this study, we investigate \algo{} across three main aspects: scenarios, methods, and hyperparameters. For scenarios, we find that \algo{} generally performs well in instruction-following and cross-lingual settings, yet struggles with mathematical reasoning. For methods, we find that classic approaches such as TA and DARE consistently achieve superior performance over alternative merging methods. We also provide a systematic and quantitative analysis of the hyperparameter tuning that these classic methods critically depend on.

% In this study, we evaluate various merging strategies to formulate comprehensive guidelines for \algo{}. Our empirical results reveal significant improvements in language and instruction-following capabilities, while also underscoring the distinct challenges associated with transferring mathematical reasoning. We find that classic methods, such as DARE, yield superior results, while tuning-free methods, such as NaN, also demonstrate robust performance. Furthermore, we analyze the hyperparameter optimization landscape to identify efficient tuning strategies, recommending Gaussian Process Bayesian Optimization (GP-BO) as the preferred method for optimizing merging hyperparameters.

% QUESTION: 感觉大家都是说结论

\end{abstract}

\section{Introduction}
\label{sec:intro}

Vision-Language Models (VLMs) have garnered increasing attention for their ability to jointly process and comprehend visual and textual information~\cite{flamingo,BLIP-2,llava}. Despite strong general performance, VLMs remain limited on specialized tasks such as visual mathematical reasoning and multilingual understanding~\cite{mathvista,cmmmu}.

Fine-tuning on specialized multimodal datasets faces significant challenges despite being a prevalent strategy to enhance domain-specific capabilities~\cite{mammothvl, wit}. First, fine-tuning VLMs requires substantial computational resources. Although Parameter-Efficient Fine-Tuning (PEFT) methods, such as LoRA and QLoRA~\cite{lora,qlora}, have effectively mitigated computational costs, acquiring sufficient high-quality training data remains a persistent bottleneck. High-quality vision datasets for specialized tasks are scarce, making it necessary to laboriously construct balanced datasets while considering factors such as data mixing ratios and domain coverage~\cite{datamix}.

Model merging offers a promising alternative for integrating expert capabilities into VLMs without extensive dataset construction or additional retraining~\cite{ta,ties,dare}. However, merging guidelines developed for homogeneous models may not apply to \algo{}. First, \algo{} is asymmetric: rather than combining peer models symmetrically, it integrates a domain-expert LLM into a VLM backbone. Second, while homogeneous merging typically aims to aggregate existing expertise, \algo{} focuses on enabling new cross-modal capabilities to emerge. For example, merging a mathematics expert with a VLM could yield visual mathematical reasoning capabilities, such as solving geometry problems, that neither model previously possessed.

Despite its potential, \algo{} remains underexplored. To establish comprehensive guidelines for \algo{}, we systematically investigate the transfer of expert LLM capabilities to VLMs from three main aspects: scenarios, methods, and hyperparameters.

\begin{figure*}[t]
    \centering
    \includegraphics[width=\linewidth]{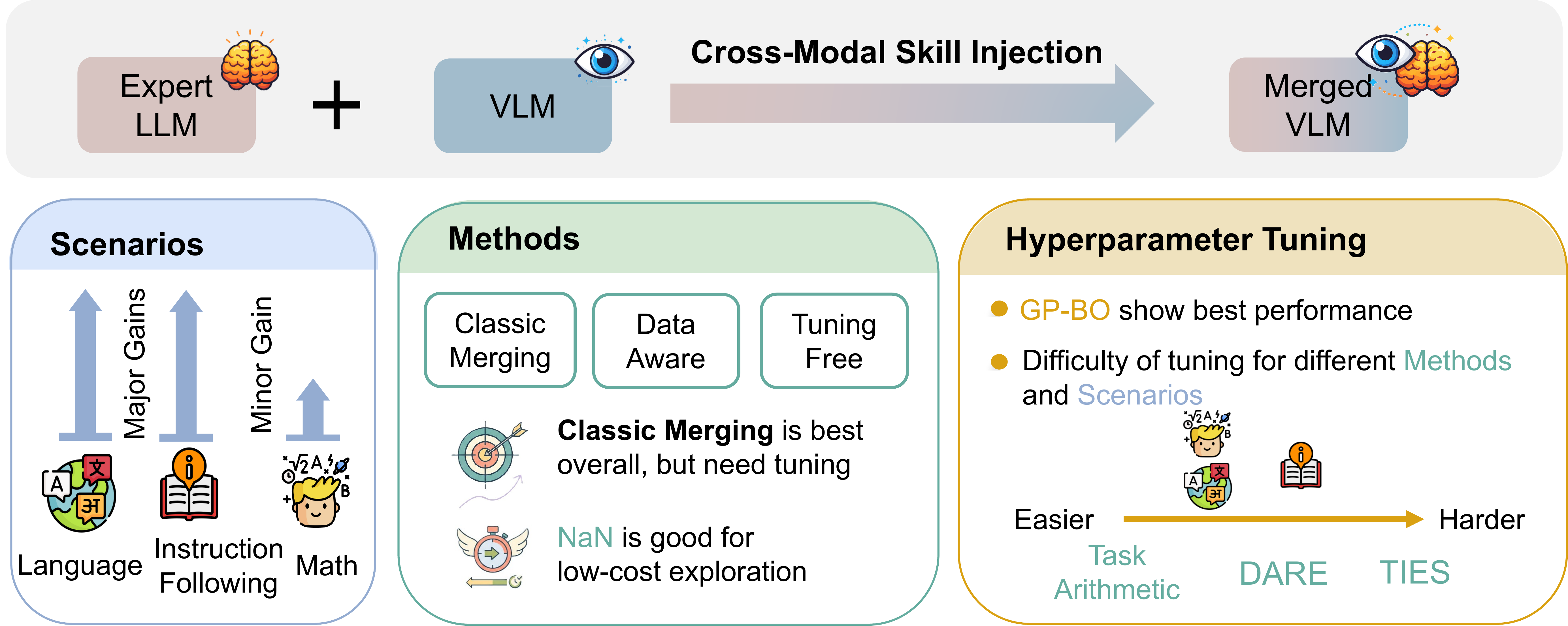}
    \caption{Overview of our work investigating \algo{} across three dimensions: \textbf{scenarios} (language ability, mathematical, instruction following), \textbf{methods} (classic, data-aware, and tuning-free merging), and \textbf{hyperparameters}. Our findings reveal that classic methods consistently outperform alternatives, language and instruction-following capabilities transfer more readily than mathematical reasoning, and we further provide quantitative analysis of the hyperparameter optimization landscape.}
    \label{fig:main}
\end{figure*}

For scenarios, we examine three distinct settings: language ability, mathematical reasoning, and instruction following. Our extensive experiments across six benchmarks demonstrate that VLMs can successfully inherit specialized capabilities from expert LLMs in language ability and instruction-following scenarios. However, we observe that transferring mathematical reasoning proves more challenging than transferring language or instruction-following abilities.

For methods, we evaluate three categories of model merging approaches. Our results show that classic merging methods consistently yield superior performance. Certain tuning-free methods also demonstrate surprisingly competitive performance; for instance, NaN~\cite{nan} achieves the second-best overall results.

For hyperparameters, since hyperparameter tuning is essential for classic methods and constitutes their primary practical cost, we conduct a thorough analysis of the hyperparameter landscape. The search space is low-dimensional but exhibits slight multimodality. Our analysis identifies GP-BO~\cite{gpbo} as the most effective optimizer, while local directional search methods, although competitive in performance, are more vulnerable to local optima.

% Additionally, we study the parameter space of the effective and ineffective merged models, providing deeper insights into how to inject new capabilities into VLMs while preserving their visual processing abilities.

% TODO: 重在指标，可以重复一遍

Our main contributions are as follows:

\begin{itemize}
    \item \textbf{Scenario Analysis.} We systematically investigate three representative scenarios for \algo{}: language ability, mathematical reasoning, and instruction following, providing insight into which scenarios are better suited to cross-modal skill transfer.
    
    \item \textbf{Method Comparison.} We conduct a comprehensive evaluation of nine merging methods across three categories (classic, data-aware, and tuning-free) on six benchmarks, offering practical guidelines for method selection based on available resources.
    
    \item \textbf{Hyperparameter Analysis.} We provide a quantitative analysis of the hyperparameter optimization landscape for classic merging methods, comparing the effectiveness of different optimization strategies.
\end{itemize}

% TODO: Add cross ref here
% In Sec 2, we analyze the the advantage and shortcoming of different merging methods, and the extent VLM can benefit from LLM capabilities in different scenarios. In Sec 3, we provide insight into the characteristics of hyperparameters in different merging methods and provide a guideline for hyperparameter tuning for \algo{}. In Sec 4, we analyze the effectiveness of \algo{} by studying the model's parameter space. 

\section{Preliminary}

% Preliminary: Problem defination 现实存在两句/benchmark 场景/模型
% ASK: should include more?

\subsection{Model Merging Methods}
\label{sec::exist_method}

% With the rapid proliferation of open-source Large Language Models (LLMs), model merging has emerged as a cost-effective strategy to combine the capabilities of multiple expert models without the need for extensive retraining or large-scale datasets. By integrating parameters from different domain experts, merging allows for the synthesis of a unified model that inherits diverse strengths. In the context of \algo{}, we categorize existing methods into three categories based on the hyperparameter tuning required and the data used: \textbf{hyperparameter-tuning-dependent methods}, \textbf{data dependent methods} and \textbf{data free methods}.

% With the rise of many open-source LLMs, LLM merging has become a prevalent way of combining different capabilities without additional training or data~\cite{modelmergingllmsmllms} and many previous model merging methods~\cite{dare,metagpt,wudi} have been developed for the merging of LLMs. 

% TA -> classic -> merging coef -> DARE, TIES -> hyperparameter cost up (mild crit) 

Model merging is closely related to the notion of task vectors~\cite{ta}, which views the parameter change induced by fine-tuning as a vector in parameter space that can encode task-specific behavior. % 先Formally, 前面拆了，refer to 最后
Formally, the task vector $\tau$ is defined as the difference between the parameters of a fine-tuned expert LLM ($\theta_{\text{finetuned}}$) and those of the base model ($\theta_{\text{base}}$):
\begin{equation}
\begin{split}
  \tau &= \theta_{\text{finetuned}} - \theta_{\text{base}} 
\end{split}
\end{equation}
Task Arithmetic constructs the merged model by adding a linear combination of these task vectors to the base model:
\begin{equation}
    \theta_{\text{merged}} = \theta_{\text{base}} + \sum_{i=1}^{n}\lambda_{i}\tau_{i}
\end{equation}
where $n$ represents the number of models being merged, and $\lambda_i$ denotes the merging coefficient for the $i$-th task vector.

% TA -> classic -> merging coef -> DARE, TIES -> hyperparameter cost up(mild crit) 
\paragraph{Classic Merging.}
The approach of linearly combining task vectors to yield a multi-skilled model is known as Task Arithmetic (TA)~\cite{ta}. In this paper, we refer to this method and its variants, such as TIES~\cite{ties} and DARE~\cite{dare}, as classic merging, characterized by combining task vectors (or their variants) with tunable merging coefficients.
In addition to model coefficients, some variants~\cite{breadcrumbs, della, mergekit} introduce a density hyperparameter to control the sparsity of task vectors. For instance, TIES-Merging~\cite{ties} prunes low-magnitude updates and resolves sign conflicts, while DARE~\cite{dare} employs random sparsification followed by rescaling. These classic merging methods typically rely on hyperparameter tuning on in-domain validation data to identify the optimal merging coefficient and, in some cases, the density parameter.

Recent works have improved upon classic merging in two directions.

\paragraph{Data-Aware Merging.}
This line of work leverages training data from each model's capability domain to refine the merging process. For instance, Fisher Merging~\cite{fisher} weights parameters based on their task-specific importance estimated from the empirical Fisher information, while RegMean~\cite{regmean} reduces differences between the intermediate representations of different models at each layer through closed-form linear regression. Intuitively, by incorporating auxiliary training data, these approaches exploit richer information and should yield better performance. However, as we will show in Section~\ref{sec::merging}, for \algo{}, the empirical gains from data-aware methods are often marginal, suggesting that, in cross-modal scenarios, practitioners can safely forgo the overhead of data collection without sacrificing much performance. % 感觉各个domain data没说清楚, escape hyperparameter tuning

\paragraph{Tuning-Free Merging.} Tuning-free methods derive merging recipes directly from model parameters, avoiding hyperparameter tuning and the need for external data. One line of work transforms task vectors through subspace operations. WUDI~\cite{wudi} exploits the observation that, in a linear layer, task vectors approximately span the corresponding input subspace, and carries out merging in this subspace. TSV~\cite{tsv} constructs layer-wise low-rank subspaces from the SVD of task matrices, and decorrelates singular directions before merging. Another line of work estimates merging coefficients directly. MetaGPT~\cite{metagpt} derives closed-form scaling coefficients under the assumption of local linearity and approximate task-vector orthogonality, while NAN~\cite{nan} estimates coefficient from inverse parameter norms. Overall, these methods are plug-and-play and appealing when data access is limited or tuning budgets are tight.

\subsection{\algoup{}}

Similar to LLM merging, expert LLMs can be merged into the language encoder backbone of a VLM to transfer specialized capabilities, which we refer to as \algo{}. 

Formally, we define \algo{} as follows: Given a VLM, denoted as $\mathcal{M}_{vlm} = ( \mathcal{E}_v, \mathcal{L}_{base}) $, where $\mathcal{E}_v$ is the vision encoder, and $\mathcal{L}_{base}$ is the LLM backbone, our goal is to endow $\mathcal{M}_{vlm}$ with domain-specific expert capabilities by integrating a domain expert LLM $\mathcal{L}_{exp}$, without modifying $\mathcal{E}_v$ or requiring full finetuning of $\mathcal{M}_{vlm}$. Specifically, we aim to construct a merged model
$$
\mathcal{M}_{merged} = (\mathcal{E}_v, f(\mathcal{L}_{base}, \mathcal{L}_{exp}))
$$
where $f(\cdot)$ denotes a merging algorithm that fuses the parameters of the backbone and expert LLMs. 
% The resulting $\mathcal{M}_{merged}$ should preserve the visual understanding of $\mathcal{M}_{vlm}$, while inheriting the specialized capabilities of $\mathcal{L}_{exp}$, and ideally exhibit emergent cross-modal abilities.

\algohead{} inherits the core advantages of LLM merging: minimal training overhead, rapid domain specialization, and independence from large-scale datasets. However, it differs fundamentally from conventional same-modality LLM merging in its functional asymmetry. In \algo{}, the VLM provides visual grounding, while the expert LLM contributes specialized capabilities. Traditional LLM merging instead combines peer models with comparable roles, primarily aiming to aggregate and preserve capabilities within a shared modality.

Rather than merely combining existing skills, \algo{} seeks to induce capabilities that neither parent model possesses in isolation. The resulting capability is emergent because the injected textual expertise must become usable under visual input, rather than remaining a purely text-side skill. Unlike same-modality merging, where the combined capabilities remain within a single representational space, cross-modal skill injection requires the interaction between visual understanding and textual expertise to arise through the merging process itself and result in a cross-modal ability.

Because \algo{} aims to induce capabilities that arise only through the interaction between expert knowledge and visual understanding, it faces challenges beyond those in conventional same-modality LLM merging. In specialized visual domains such as medical imaging, legal document analysis, and scientific figure understanding, paired image-text data for validation is often scarce or expensive to collect, which imposes tighter constraints on hyperparameter tuning. Moreover, merging methods designed to reconcile conflicts among peer models may be less effective in this setting, because the LLM and VLM play asymmetric roles rather than contributing comparable capabilities within a shared modality.

\section{Scenario and Methods for \algoup{}}
\label{sec::merging}

\begin{table*}[htbp]
\centering
\footnotesize
\setlength{\tabcolsep}{2pt}
\renewcommand{\arraystretch}{1.2}

\begin{tabularx}{\textwidth}{
  >{\raggedright\arraybackslash}l|
  >{\centering\arraybackslash}X 
  >{\centering\arraybackslash}X||
  >{\centering\arraybackslash}X
  >{\centering\arraybackslash}X|
  >{\centering\arraybackslash}X
  >{\centering\arraybackslash}X||
  >{\centering\arraybackslash}X
  >{\centering\arraybackslash}X|
  >{\centering\arraybackslash}X
  >{\centering\arraybackslash}X|
  c
}
% \hline
\toprule
\multirow{2}{*}{\makecell[c]{\textbf{Method}}} & 
\textbf{CMMMU} & \textbf{JMMMU} &
\multicolumn{2}{c|}{\textbf{MathVista }\footnotesize{(math)}} &
\multicolumn{2}{c||}{\textbf{MathVerse}} &
\multicolumn{2}{c|}{\textbf{MIA-Bench}} &
\multicolumn{2}{c|}{\textbf{WildVision}} & 
\multirow{2}{*}{\textbf{Avg}} \\
\cmidrule{2-11}

& \textbf{Mistral} & \textbf{LLaMA}
& \textbf{Mistral} & \textbf{LLaMA}
& \textbf{Mistral} & \textbf{LLaMA}
& \textbf{Qwen2} & \textbf{Idefics2}
& \textbf{Qwen2} & \textbf{Idefics2} & \\
\midrule

TA       & 25.0\textcolor{green!50!black}{\scriptsize $\uparrow$2.4} & \textbf{42.1}\textcolor{green!50!black}{\scriptsize $\uparrow$5.7} & 26.7\textcolor{green!50!black}{\scriptsize $\uparrow$1.4} & 29.2\textcolor{green!50!black}{\scriptsize $\uparrow$3.4} & 17.1\textcolor{green!50!black}{\scriptsize $\uparrow$1.9} & 15.0\textcolor{red}{\scriptsize $\downarrow$1.1} & \textbf{76.1}\textcolor{green!50!black}{\scriptsize $\uparrow$43.1} & 75.9\textcolor{green!50!black}{\scriptsize $\uparrow$66.1} & \textbf{50.1}\textcolor{green!50!black}{\scriptsize $\uparrow$34.0} & 22.8\textcolor{green!50!black}{\scriptsize $\uparrow$19.0} & \textbf{38.0} \\
DARE     & \textbf{28.5}\textcolor{green!50!black}{\scriptsize $\uparrow$5.9} & 39.0\textcolor{green!50!black}{\scriptsize $\uparrow$2.6} & \textbf{29.0}\textcolor{green!50!black}{\scriptsize $\uparrow$3.7} & \underline{30.6}\textcolor{green!50!black}{\scriptsize $\uparrow$4.8} & 16.1\textcolor{green!50!black}{\scriptsize $\uparrow$0.9} & \textbf{17.0}\textcolor{green!50!black}{\scriptsize $\uparrow$0.9} & \underline{75.4}\textcolor{green!50!black}{\scriptsize $\uparrow$42.4} & 74.3\textcolor{green!50!black}{\scriptsize $\uparrow$64.5} & 47.5\textcolor{green!50!black}{\scriptsize $\uparrow$31.4} & 19.9\textcolor{green!50!black}{\scriptsize $\uparrow$16.1} & \underline{37.7} \\
TIES     & 27.4\textcolor{green!50!black}{\scriptsize $\uparrow$4.8} & \underline{40.9}\textcolor{green!50!black}{\scriptsize $\uparrow$4.5} & 23.7\textcolor{red}{\scriptsize $\downarrow$1.6} & 27.1\textcolor{green!50!black}{\scriptsize $\uparrow$1.3} & 14.5\textcolor{red}{\scriptsize $\downarrow$0.7} & 15.3\textcolor{red}{\scriptsize $\downarrow$0.8} & 74.7\textcolor{green!50!black}{\scriptsize $\uparrow$41.7} & \underline{76.3}\textcolor{green!50!black}{\scriptsize $\uparrow$66.5} & 45.1\textcolor{green!50!black}{\scriptsize $\uparrow$29.0} & \textbf{23.3}\textcolor{green!50!black}{\scriptsize $\uparrow$19.5} & 36.8 \\
Fisher   & 25.3\textcolor{green!50!black}{\scriptsize $\uparrow$2.7} & 36.0\textcolor{red}{\scriptsize $\downarrow$0.4} & 22.3\textcolor{red}{\scriptsize $\downarrow$3.0} & \textbf{32.4}\textcolor{green!50!black}{\scriptsize $\uparrow$6.6} & 15.9\textcolor{green!50!black}{\scriptsize $\uparrow$0.7} & 13.9\textcolor{red}{\scriptsize $\downarrow$2.2} & 68.6\textcolor{green!50!black}{\scriptsize $\uparrow$35.6} & \textbf{78.3}\textcolor{green!50!black}{\scriptsize $\uparrow$68.5} & 23.1\textcolor{green!50!black}{\scriptsize $\uparrow$7.0} & \textbf{23.3}\textcolor{green!50!black}{\scriptsize $\uparrow$19.5} & 33.9 \\
RegMean  & \underline{28.1}\textcolor{green!50!black}{\scriptsize $\uparrow$5.5} & 40.5\textcolor{green!50!black}{\scriptsize $\uparrow$4.1} & \underline{27.1}\textcolor{green!50!black}{\scriptsize $\uparrow$1.8} & 26.0\textcolor{green!50!black}{\scriptsize $\uparrow$0.2} & \textbf{17.9}\textcolor{green!50!black}{\scriptsize $\uparrow$2.7} & 16.0\textcolor{red}{\scriptsize $\downarrow$0.1} & 69.1\textcolor{green!50!black}{\scriptsize $\uparrow$36.1} & 4.9\textcolor{red}{\scriptsize $\downarrow$4.9}  & 26.4\textcolor{green!50!black}{\scriptsize $\uparrow$10.3} & 2.3\textcolor{red}{\scriptsize $\downarrow$1.5} & 25.8 \\
WUDI     & 25.4\textcolor{green!50!black}{\scriptsize $\uparrow$2.8} & 35.6\textcolor{red}{\scriptsize $\downarrow$0.8} & 24.8\textcolor{red}{\scriptsize $\downarrow$0.5} & 29.4\textcolor{green!50!black}{\scriptsize $\uparrow$3.6} & 11.3\textcolor{red}{\scriptsize $\downarrow$3.9} & \underline{16.2}\textcolor{green!50!black}{\scriptsize $\uparrow$0.1} & 66.8\textcolor{green!50!black}{\scriptsize $\uparrow$33.8} & 5.3\textcolor{red}{\scriptsize $\downarrow$4.5}  & 25.4\textcolor{green!50!black}{\scriptsize $\uparrow$9.3} & 3.1\textcolor{red}{\scriptsize $\downarrow$0.7} & 24.3 \\
TSV      & 24.0\textcolor{green!50!black}{\scriptsize $\uparrow$1.4} & 38.3\textcolor{green!50!black}{\scriptsize $\uparrow$1.9} & 23.5\textcolor{red}{\scriptsize $\downarrow$1.8} & 29.7\textcolor{green!50!black}{\scriptsize $\uparrow$3.9} & 16.0\textcolor{green!50!black}{\scriptsize $\uparrow$0.8} & 15.3\textcolor{red}{\scriptsize $\downarrow$0.8} & 54.8\textcolor{green!50!black}{\scriptsize $\uparrow$21.8} & 72.4\textcolor{green!50!black}{\scriptsize $\uparrow$62.6} & 15.2\textcolor{red}{\scriptsize $\downarrow$0.9} & 16.5\textcolor{green!50!black}{\scriptsize $\uparrow$12.7} & 29.9 \\
MetaGPT  & \underline{28.1}\textcolor{green!50!black}{\scriptsize $\uparrow$5.5} & 37.1\textcolor{green!50!black}{\scriptsize $\uparrow$0.7} & 18.4\textcolor{red}{\scriptsize $\downarrow$6.9} & 25.3\textcolor{red}{\scriptsize $\downarrow$0.5} & 12.3\textcolor{red}{\scriptsize $\downarrow$2.9} & 15.4\textcolor{red}{\scriptsize $\downarrow$0.7} & 60.6\textcolor{green!50!black}{\scriptsize $\uparrow$27.6} & 76.1\textcolor{green!50!black}{\scriptsize $\uparrow$66.3} & 36.3\textcolor{green!50!black}{\scriptsize $\uparrow$20.2} & \underline{22.9}\textcolor{green!50!black}{\scriptsize $\uparrow$19.1} & 33.3 \\
NaN      & 26.5\textcolor{green!50!black}{\scriptsize $\uparrow$3.9} & 40.5\textcolor{green!50!black}{\scriptsize $\uparrow$4.1} & 23.7\textcolor{red}{\scriptsize $\downarrow$1.6} & 29.4\textcolor{green!50!black}{\scriptsize $\uparrow$3.6} & \underline{17.3}\textcolor{green!50!black}{\scriptsize $\uparrow$2.1} & \textbf{17.0}\textcolor{green!50!black}{\scriptsize $\uparrow$0.9} & 73.8\textcolor{green!50!black}{\scriptsize $\uparrow$40.8} & 73.0\textcolor{green!50!black}{\scriptsize $\uparrow$63.2} & \underline{47.8}\textcolor{green!50!black}{\scriptsize $\uparrow$31.7} & 21.5\textcolor{green!50!black}{\scriptsize $\uparrow$17.7} & 37.1 \\
\midrule
Base VLM    & 22.6 & 36.4 & 25.3 & 25.8 & 15.2 & 16.1 & 33.0 & 9.8  & 16.1 & 3.8 & 20.4 \\
\bottomrule
\end{tabularx}

\caption{Performance comparison across merging methods, datasets, and models, with average scores computed for each method. Colored deltas indicate improvement (\textcolor{green!50!black}{green}) or degradation (\textcolor{red}{red}) relative to the base VLM. \textbf{Bold}: best; \underline{underline}: second best. Key observations: (1) language and instruction-following abilities show substantial improvements, while mathematical reasoning remains challenging; (2) classic methods (TA, DARE) achieve the best overall performance, and NaN offers a competitive tuning-free alternative.}
\label{tab:benchmark_results}
\end{table*}

% and display the average for each method across datasets. 1) Language and Instruction following has greater improvement while math is chanllenging. 2) classic methods such as DARE, TA is generally best overall and NaN as a tuning free method have good performance as well. Although it should be noted that despite better average
% 按照价格

In this section, we derive practical guidelines regarding which scenarios and merging strategies are more suitable for \algo{}. Specifically, we focus on three representative visual capability scenarios: language ability, mathematical reasoning ability, and instruction-following ability. These are among the most commonly evaluated and practically relevant abilities in prior work~\cite{scenario_work1, scenario_work2}.

\paragraph{Model Settings.}
We conduct experiments using publicly available models from Hugging Face. Our VLM cover a diverse set of architectures, including Idefics2~\cite{idefics2}, LLaVA with both Mistral and LLaMA backbones~\cite{llava_model, improvedllava, llavanext}, and Qwen2-VL~\cite{qwen2vl}. For each scenario, we pair the VLM with a domain-expert LLM that shares the same base architecture. Specifically, for language understanding, we use Mistral-7B-v0.3-Chinese-Chat and Llama-3-ELYZA-JP-8B to inject Chinese and Japanese capabilities, respectively. For mathematical reasoning, we use the DART-Math series built on LLaMA and Mistral backbones. For instruction following, we pair Qwen2-VL with Qwen2-Instruct and Idefics2-base with Mistral-7B-Instruct. Appendix~\ref{sec:checkpoint} lists the full checkpoint names and pairing.

\paragraph{Evaluation Settings.}
We employ six benchmarks to evaluate the merged VLMs under different scenarios.
% To assess the merged VLMs under different scenarios, we employ six benchmarks.
For visual mathematical reasoning, we use MathVista (math subset)~\cite{mathvista} and MathVerse~\cite{mathverse};
for language understanding with visual inputs, CMMMU (Chinese)~\cite{cmmmu} and JMMMU (Japanese)~\cite{jmmmu};
and for visual instruction following, MIA-Bench~\cite{miabench} and WildVision~\cite{wildvision}.
For each benchmark, 20\% of the samples are randomly held out for hyperparameter tuning, while the remaining 80\% are used exclusively for testing.

To estimate the cost of merging and hyperparameter tuning, we conduct experiments on a single A800 GPU, utilizing the Python package MergeKit~\cite{mergekit} for merging and lmms-eval~\cite{lmmsevalrealitycheckevaluation,lmmseval} for evaluation. The total cost is computed by summing the GPU cost and the OpenAI API cost for evaluating the merged models. The reported total duration includes both merging and evaluation time.

\paragraph{Methods Settings.}
We employ the three categories of merging methods introduced in Section~\ref{sec::exist_method}; further details on hyperparameter settings and calibration data are provided in the Appendix~\ref{sec:data}.
(1) \textit{Classic merging methods} (Task Arithmetic, DARE, TIES-Merging): We perform grid search over VLM and LLM coefficients in $\{0.1, 0.3, 0.5, 0.7, 0.9\}$; for methods requiring a density parameter, we additionally search over $\{0.2, 0.4, 0.6, 0.8\}$.
(2) \textit{Data-aware methods} (Fisher, RegMean): We use 500 calibration samples drawn from corresponding domain-specific datasets. Specifically, the VLM is calibrated on a vision-instruction dataset (LLaVA-Instruct-150K)~\citep{llava_model, improvedllava, llavanext}, while the expert LLM is calibrated on scenario-matched text data: Alpaca-Zh for Chinese~\citep{alpaca_zh}, Japanese-Alpaca-Data for Japanese~\citep{alpaca_ja}, Competition-Math mixed with GSM8K for mathematical reasoning~\citep{math500, gsm8k}, and Dolly-15K for instruction following~\citep{dolly}. 
(3) \textit{Tuning-free methods} (WUDI, TSV, MetaGPT, NaN): These methods can be applied without additional hyperparameter tuning or reliance on external data. % TODO: methods 要再引用一遍吗？

% To conduct out experiment, for each scenario, we selected two benchmarks: for lingual ability we used Chinese and Japanese ability and chose CMMMU and JMMMU respectively; for mathematical ability, we chose two datasets Mathvista and MathVerse; for instruction following ability, we select miabench and wildvision. For models, we selected VLMs from the Qwen, Idefics and Llava family, and corresponding LLMs of Qwen, Llama and Mistral structure with expert ability. 

% We use the merging methods mentioned above: for hyperparameter-tuning dependent methods, we select Task Arithmetic, Darelinear and TIES; for data-dependent methods, we select Fisher and Regmean and for data-free methods, we choose WUDI-merging, MetaGPT and NaN. Because we need domain visual data for hyperparameter-tuning dependent models, we select 20\% of the benchmark itself for hyperparameter-tuning and used the 80\% left for testing. For data-dependent methods, we need domain text data and generic visual data, and we leave the specifics to the Appendix. And the results of our experiments is shown as follows.

% For language ability transfer, the merged models demonstrate notable improvements of 4.2 and 2.6 absolute points for Chinese and Japanese, respectively. 

\paragraph{Finding 1: \algohead{} is more effective for transferring language and instruction-following abilities than for transferring reasoning and mathematical abilities.}

As shown in Table~\ref{tab:benchmark_results}, the most substantial improvements are observed in language and instruction-following benchmarks, with average gains of 3.4 and 28.1 absolute points, respectively. Instruction following is particularly favorable for merging, as such capabilities are largely modality-agnostic, enabling effective transfer to base VLMs. Notably, merging Idefics2-base with an instruction-following LLM expert achieves over 70 on MIA-Bench, surpassing the fine-tuned Idefics2 (56.4).

In contrast, mathematical ability transfer yields much smaller and less consistent gains. On MathVista, no significant improvement is observed on the full benchmark (see Appendix~\ref{sec:mathvista}); even on the math-specific subset, the average gain is only 1.05 absolute points. On MathVerse, more than half of the merging configurations are detrimental. Overall, DARE is the only method that yields consistent improvements in mathematical scenarios, while more than 40\% of merged models are unable to outperform the original VLM.

One plausible explanation of less effective merging outcomes in math scenario is that visual mathematical reasoning is a highly entangled capability, demanding simultaneous coordination of visual perception (e.g., reading diagrams) and multi-step logical reasoning. Such entwinement may be difficult to reconstruct through simple parameter-space interpolation.
Moreover, although some mathematical reasoning skills do transfer, these gains are often offset by degraded visual understanding. This interpretation is supported by the full MathVista breakdown (see Appendix~\ref{sec:mathvista}): merging with a math-finetuned LLM often modestly improves the ``Math'' subset while hurting the more perception-heavy ``General'' subset. Concretely, merged models become better at reasoning-heavy questions such as ``Find the length of AC in the isosceles triangle ABC,'' yet lose accuracy on visually dependent questions such as ``Does Aqua have the minimum area under the curve?''

% MathVista form
% MathVerse form

\paragraph{Finding 2: Classic merging methods consistently outperform other approaches with superior stability. NaN offers a viable low-cost option for preliminary exploration, though with modest performance trade-offs.} % degradation

Among all merging methods, classic merging yield the strongest and most stable results (Table~\ref{tab:benchmark_results}). Task Arithmetic outperforms TIES on average, while DARE demonstrates the best consistency, being the only method that yields consistent improvements across all scenarios. These methods do require domain-specific visual-text data and hyperparameter tuning (Table~\ref{tab:cost}), yet this investment consistently translates into superior performance.

For data-aware merging methods, RegMean achieves better results than Fisher while also being more efficient, as it requires only activation information without gradient computation. RegMean even achieves the highest accuracy among all methods on the MathVerse benchmark, though its overall average still falls short of classic merging methods and NaN.

Among tuning-free methods, which require neither additional data nor hyperparameter tuning, NaN stands out as the most effective. While its performance is slightly inferior and less stable compared to classic methods, NaN serves as a practical low-barrier entry point, enabling practitioners to quickly gauge whether model merging is promising for a new scenario before committing to more resource-intensive hyperparameter search. Moreover, NaN is among the most efficient approaches: along with MetaGPT, it is the least time-intensive and consumes only a fraction of WUDI and TSV's total computing time while surpassing their accuracy (see Appendix~\ref{sec:efficiency}). This combination of minimal overhead and reasonable performance makes NaN particularly well suited for rapid feasibility assessment in unfamiliar domains.

In summary, classic merging methods remain the gold standard when accuracy and stability are paramount, provided that domain visual data is available and hyperparameter tuning is feasible. For scenarios requiring quick preliminary exploration with minimal overhead, NaN offers a reasonable first-pass solution to assess merging potential before deeper investment. When domain text data is accessible, RegMean is preferable to Fisher, though its results may not match classic methods or NaN despite the additional cost.

\begin{table}[htbp]
    \centering
    \small
    \begin{tabular}{lcc}
        \toprule
        Dataset & Total cost (dollars) & Total Duration \\
        \midrule
        CMMMU     & 2040.21 & 29h 36m 17s \\
        JMMMU     & 1976.88 & 29h 43m 15s \\
        MathVista & 1644.35 & 24h 13m 26s \\
        MathVerse & 3599.98 & 45h 49m 2s  \\
        Miabench  & 2008.87 & 19h 14m 30s \\
        Wildvision& 8044.11 & 45h 0m 22s  \\
        \bottomrule
    \end{tabular}
    \caption{Average cost and duration of hyperparameter tuning for DARE on each dataset using an A800 GPU, highlighting the substantial computational overhead of the optimization process.}
    \label{tab:cost}
\end{table}

% \section{Efficient Hyperparameter Tuning}
\section{Analysis of Hyperparameter Optimization Landscape and Strategies}

As discussed in Section~\ref{sec::merging}, hyperparameter-tuning merging methods consistently achieve strong results, but they can be time-consuming and resource-intensive. Therefore, it is essential to provide guidelines for effective hyperparameter tuning.

% TODO add more 结论 here

\subsection{Revisiting the Sum-to-One Constraint} % Validity

Existing approaches often constrain the sum of merging coefficients to 1~\cite{bringreasontovision}. However, this constraint is largely intuitive and lacks empirical validation. Our experiments reveal that while the sum-to-one constraint proves effective in most scenarios, it can lead to substantial performance degradation in others.

Let $S := \lambda_{\text{VLM}} + \lambda_{\text{LLM}}$ denote the sum of merging coefficients. To investigate whether restricting the search space around $S = 1$ suffices to identify the optimal hyperparameters, we introduce the notion of \textit{relative regret}. Specifically, we first identify the global optimum over the full coefficient grid obtained in Section~\ref{sec::merging}, then compute the best achievable performance when the search is restricted to $S = 1$ or $S \in [0.8, 1.2]$. The relative regret is defined as the percentage of performance degradation relative to the global optimum caused by the restricted search space. % TODO: cross ref？

As shown in Table~\ref{tab:regret}, restricting the search to $S \approx 1$ can be suboptimal. In certain scenarios such as WildVision and MathVerse, the $S = 1$ constraint leads to substantial performance degradation, with relative regret as high as 25\%, and relaxing $S$ to $[0.8, 1.2]$ provides only limited mitigation. While most other cases exhibit regret within 5\% under the $S = 1$ constraint, and often drop to zero when the search space is relaxed to $S \in [0.8, 1.2]$, the existence of such high-regret cases demonstrates the need for a broader search space.

\begin{table}[t]
\centering
\small
\setlength{\tabcolsep}{3pt}
\begin{tabular}{l|cc|cc|cc}
\toprule
\multirow{2}{*}{\textbf{Scenario}} & \multicolumn{2}{c|}{\textbf{TA}} 
                  & \multicolumn{2}{c|}{\textbf{DARE}} 
                  & \multicolumn{2}{c}{\textbf{TIES}} \\
 & {$S{=}1$} & {$S{\approx}1$} 
 & {$S{=}1$} & {$S{\approx}1$} 
 & {$S{=}1$} & {$S{\approx}1$}  \\
\midrule
CMMMU       & 0     & 0     & 0     & 0     & 0     & 0     \\
JMMMU       & 3.6   & 0     & 3.6   & 0     & 0.9   & 0.9   \\
MVis-L      & 0     & 0     & 0     & 0     & 0     & 0     \\
MVis-M      & 0     & 0     & 2.6   & 2.6   & 0     & 0     \\
MV-L3       & 5.3   & 0     & 7.8   & 2.6   & 1.5   & 0     \\
MV-M        & 2.7   & 0     & 13    & 0     & 4.5   & 4.5   \\
MIA-I2      & 1.5   & 1.5   & 3.3   & 1.3   & 0     & 0     \\
MIA-Q2      & 1.2   & 0     & 0     & 0     & 1.9   & 0     \\
WV-I2       & 25    & 0     & 19    & 8.1   & 13    & 13    \\
WV-Q2       & 0     & 0     & 5.7   & 0     & 13    & 0     \\
\bottomrule
\end{tabular}
\caption{Relative regret (\%) under constrained search spaces. Restricting the coefficient sum $S$ to approximately 1 leads to substantial performance degradation in certain scenarios (e.g., WildVision). $S{\approx}1$: $S \in [0.8, 1.2]$. Abbreviations—MV: MathVerse, MVis: MathVista, MIA: MIA-Bench, WV: WildVision; L3: LLaMA3, M: Mistral, I2: Idefics2, Q2: Qwen2.}
\label{tab:regret}
\end{table}

\subsection{Benchmarking Hyperparameter Optimization Algorithms}

The demand for comprehensive hyperparameter search is significant, particularly when the optimal merging coefficients do not adhere to the $S=1$ constraint. Given that the evaluation cost of merged models is substantial, employing a sample-efficient hyperparameter optimization algorithm is crucial to minimize the number of trials. In this section, we evaluate the performance of various hyperparameter optimization algorithms in the context of \algo{}.

\paragraph{Optimization Algorithms.}

To ensure a comprehensive evaluation of the optimization landscape, we consider a diverse spectrum of derivative-free algorithms. We include \textit{Random Search} and \textit{Sobol Sequences}~\cite{sobol}, the latter providing low-discrepancy quasi-random coverage, as reference methods. Beyond these, we evaluate \textit{CMA-ES}~\cite{cmaes}, which adapts its sampling covariance during evolution, and \textit{GP-BO}~\cite{gpbo}, which fits a Gaussian process surrogate to guide the search. We also examine \textbf{direct-search methods}~\cite{directsearch}, including \textit{Pattern Search}~\cite{pattern}, a coordinate-descent variant, and \textit{Powell's Method}~\cite{powell}, which relies on sequential line searches. Comparing these local direct-search methods with global optimizers (e.g., GP-BO) allows us to more thoroughly examine whether the hyperparameter landscape contains multiple local optima: a substantial performance gap would suggest that greedy search methods can easily become trapped in suboptimal regions of the search space.

\paragraph{Evaluation Metric.}

% \begin{figure}[t]
%     \centering
%     \includegraphics[width=0.6\linewidth]{image/task_arithmetic_val_test_regret.png}
%     \caption{The gap between validation and test sets is small.}
%     \label{fig:valtest}
% \end{figure}

We measure tuning performance using the \emph{normalized regret} after $k$ evaluations, defined as
\[
r_k \;=\; 
\frac{g_{\max} - g^\star(k)}
     {\,g_{\max} - g_{\min}\,},
\]
where $g_{\max}$ and $g_{\min}$ denote the global maximum and minimum of the objective function, respectively, and $g^\star(k)$ is the best objective value observed within the first $k$ evaluations. The normalized regret satisfies $r_k \in [0,1]$, with $r_k = 0$ indicating that the global optimum has been found.

To compare the effectiveness of different optimization algorithms, we report \emph{regret-over-random} (RoR), defined as
\[
\mathrm{RoR}_k \;=\; \frac{r_k^{\text{method}}}{r_k^{\text{random}}},
\]
which normalizes each method's regret by that of Random Search. This ratio eliminates the influence of varying optimization difficulty across datasets, allowing for direct comparison: $\mathrm{RoR}_k < 1$ indicates the method outperforms Random Search, while $\mathrm{RoR}_k > 1$ indicates underperformance.

\begin{figure}[t]
    \centering
    \includegraphics[width=\linewidth]{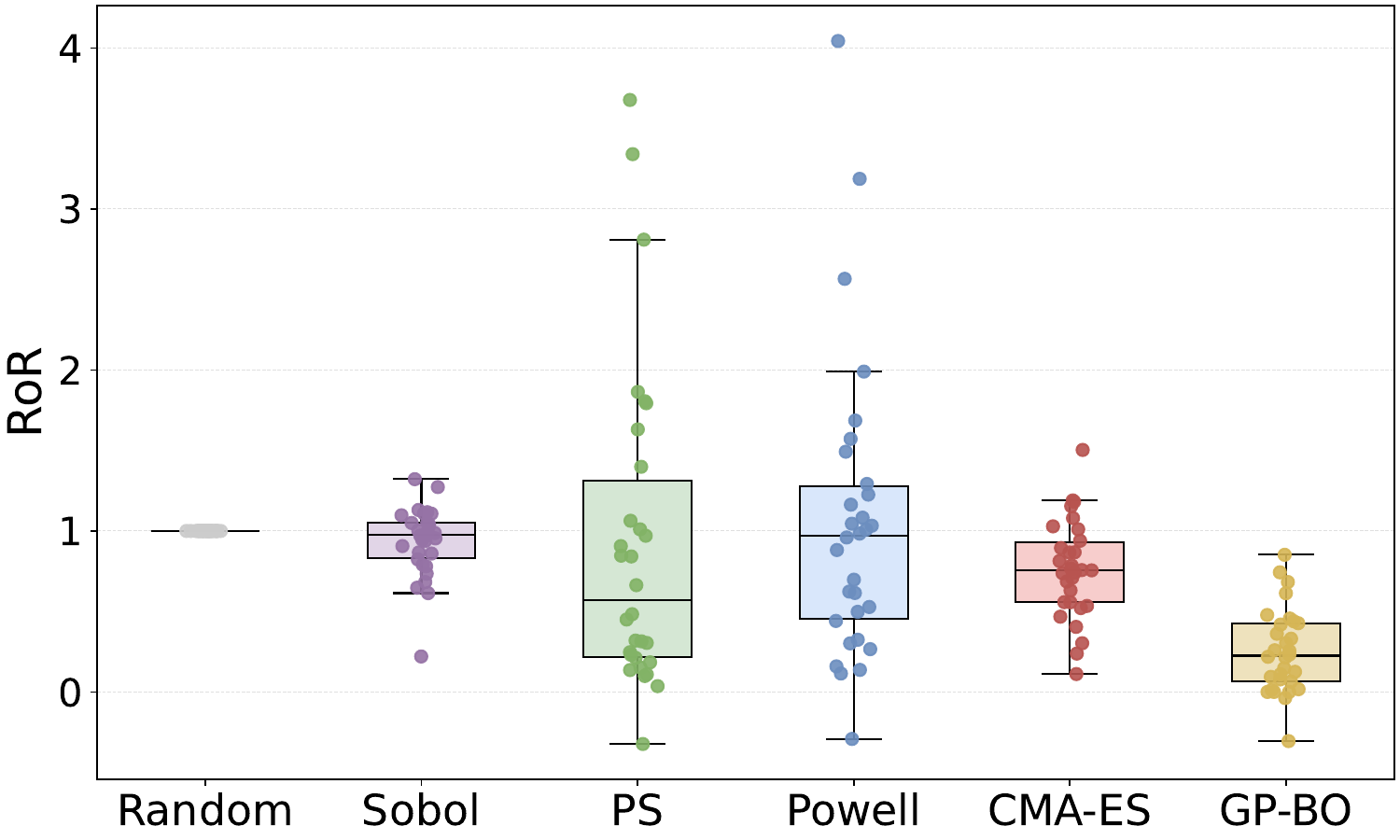}
    \caption{Distribution of regret-over-random (RoR) across optimization algorithms. Values below 1 indicate better performance than random search. GP-BO achieves the lowest median RoR, demonstrating superior sample efficiency. Pattern Search shows competitive mean performance but also high variance across runs due to sensitivity to initialization.}
    \label{fig:difficulty}
\end{figure}

\paragraph{Experimental Setup.}

We set the evaluation budget $k$ to balance computational cost and search effectiveness: $k = 40$ for Task Arithmetic, which involves two hyperparameters, and $k = 60$ for TIES and DARE, which involve three. All optimizers employ a multi-start strategy, restarting from a new random initial point upon convergence or local budget exhaustion until the total budget is depleted. To ensure fair comparison and reduce variance from random initialization, we conduct 10 independent runs per optimizer with different random seeds and report averaged results.

All hyperparameter searches are conducted on the validation set. The optimization trajectories on the validation and test sets exhibit strong agreement, with normalized regret decreasing consistently on both as the search progresses. (See Appendix~\ref{sec:valtest}) This close correspondence justifies our use of the validation set as a low-cost proxy, avoiding the prohibitive expense of repeated test-set evaluations while preserving the reliability of our conclusions.

\begin{figure}[t]
    \centering
    \includegraphics[width=\linewidth]{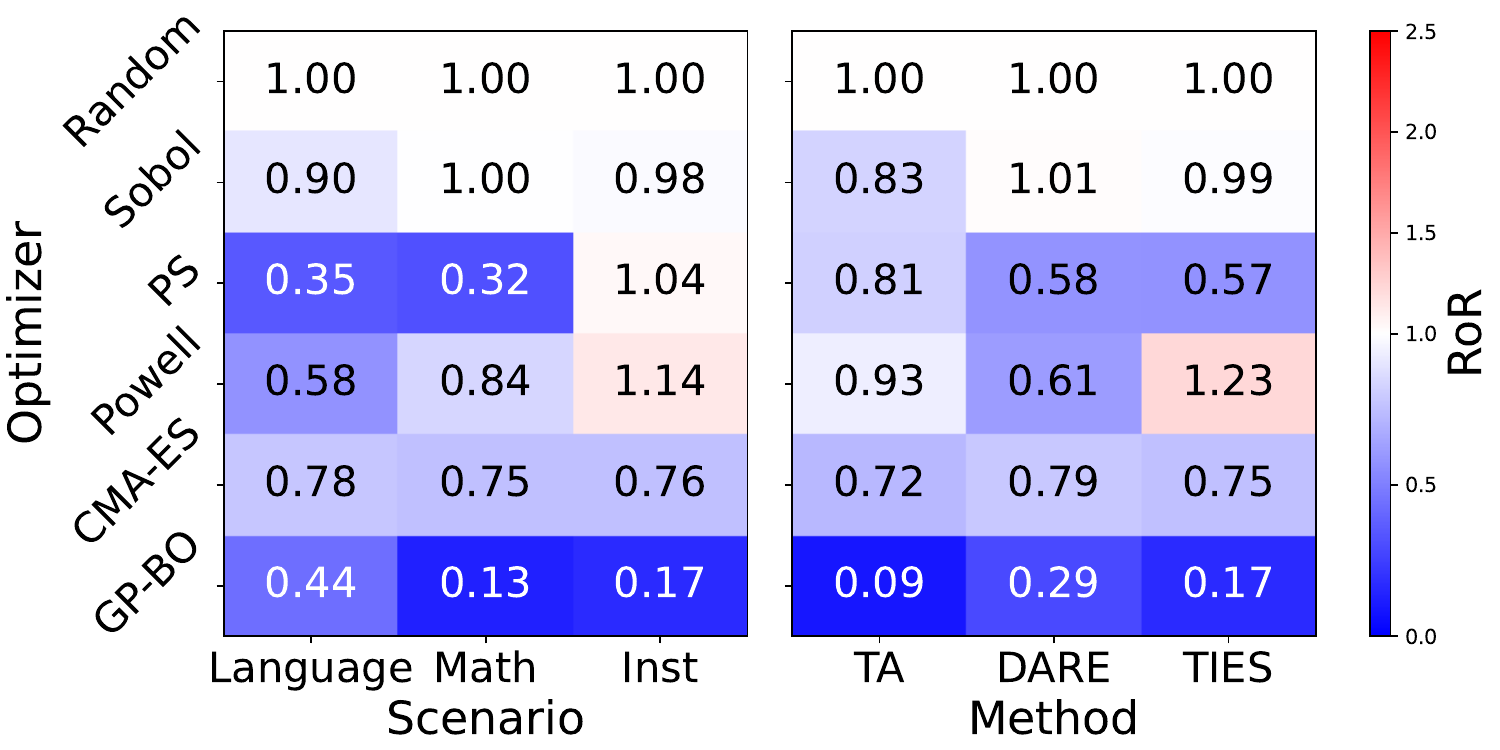}
    \caption{Regret-over-random (RoR) heatmap across optimization algorithms, scenarios and merging methods. Lower values (\textcolor{blue}{blue}) indicate better performance. Instruction-following tasks show larger gaps between local and global methods.}
    \label{fig:heatmap}
\end{figure}

% To ensure a comprehensive evaluation of the optimization landscape, 
% we consider a diverse spectrum of derivative-free algorithms. 
% As \textbf{global sampling baselines}, we include Random Search 
% and quasi-random Sobol Sequences. \textbf{Model-based solvers}, 
% namely CMA-ES and GP-BO, are employed to handle non-convexity, 
% multimodality, and sensitivity to initialization. 
% \textbf{Direct-search methods}, including Pattern Search and 
% Powell's Method, provide local perspectives on basin geometry 
% and conditioning.

% Due to the fact that obtaining the objective, evaluation the merged model's outputs, is expensive and time consuming, we 
% To account for dataset-dependent difficulty, we normalize each method’s AURC by the AURC of Random Search on the same dataset, producing a comparable difficulty-invariant score:
% \begin{equation}
%     \mathrm{nAURC}_{m,d} =
%     \frac{\mathrm{AURC}_{m,d}}{\mathrm{AURC}_{\text{Random},d}}.
% \end{equation}

\subsubsection{Comparison of Optimization Algorithms}

Among all methods evaluated, Gaussian Process Bayesian Optimization (GP-BO) consistently achieves the lowest median regret-over-random (Figure~\ref{fig:difficulty}), showing the strongest sample efficiency for model merging across our evaluations.

Pattern Search attains competitive mean performance but exhibits high variance across runs. As a local search method that greedily descends along coordinate directions, it is prone to converging to local optima, and its sensitivity to initialization reflects the presence of multiple local optima in the hyperparameter landscape. Nevertheless, its strong average performance suggests that while the objective surface is non-convex, the number of local optima remains limited and their quality is relatively high, rendering the space reasonably tractable despite lacking a unique global optimum.

CMA-ES shows limited effectiveness in our experiments, as it is designed for higher-dimensional problems with larger evaluation budgets. In our low-dimensional setting with only 2–3 hyperparameters and 40–60 evaluations, CMA-ES cannot complete enough generations for its covariance matrix adaptation to converge, and its population-based sampling is less efficient than the sequential, model-guided search used by GP-BO. Powell's method, another gradient-free local optimizer, shows performance broadly comparable to that of Pattern Search, but its overall results are slightly weaker.

% Sobol sequence and random search, lacking any adaptive mechanism, serve primarily as baselines and unsurprisingly yield the weakest results.

\subsubsection{Optimization Landscape Across Tasks and Merging Methods}

% \begin{figure}[t]
%     \centering
%     \includegraphics[width=\linewidth]{image/boxplot_method_regret.pdf}
%     \caption{Difficulty of hyperparameter tuning for different merging methds}
%     \label{fig:methods}
% \end{figure}

% 换个用语

We further analyze how optimization difficulty varies with the combination of merging method and target task, as shown in Figure~\ref{fig:heatmap}.

\paragraph{Across Task Domains.}

Instruction-following tasks are harder to optimize with local methods than mathematical and language tasks. For mathematical and language tasks, local methods (Pattern Search, Powell) achieve performance comparable to global optimizers (GP-BO, CMA-ES). In contrast, for instruction-following tasks, local methods suffer significant degradation while global methods remain unaffected, suggesting the presence of local optima that trap greedy searches.

\paragraph{Across Merging Methods.}

Among the three merging methods, Task Arithmetic is the easiest to optimize, followed by DARE, with TIES being slightly more challenging. Task Arithmetic benefits from a two-dimensional search space that requires fewer optimization rounds, making it straightforward to optimize in absolute terms, despite some degradation observed with local methods. DARE, although involving three hyperparameters, exhibits a smooth optimization landscape where local methods still achieve competitive performance. TIES presents somewhat more difficulty, with Powell's method suffers notable performance degradation.

% \subsubsection{Optimization Landscape Across Tasks and Merging Methods}

% We further investigate the impact of different methods on search difficulty, as shown in Figure~\ref{fig:difficulty}. The heatmap reveals distinct patterns across domains. Mathematical and Instruction Following tasks appear to be more "searchable," with GP-BO achieving impressive normalized regrets of 0.10 and 0.20, respectively. This implies that the hyperparameter landscapes in these domains likely exhibit smooth or convex characteristics that facilitate efficient optimization. In contrast, Language tasks present a greater challenge, with the best optimizer (GP-BO) only reducing the regret to 0.52. This suggests that the hyperparameter sensitivity in multilingual scenarios might be more complex or less predictable, diminishing the relative advantage of advanced search strategies over random sampling.

% Notably, all local search methods (Pattern Search, Powell, and Nelder-Mead) exhibit degraded performance on TIES compared to Task Arithmetic and DARE, indicating that its landscape has local optima. In contrast, global optimization methods such as CMA-ES and GP-BO maintain consistent performance across all three merging methods, suggesting that the increased difficulty of TIES stems specifically from its multimodal structure rather than from a harder-to-locate global optimum as it traps greedy local searches but pose little challenge to global strategies.

\section{Related Work} % 这段之后会改, 之后改
% ASK: at the back?
% 分析性工作 MergeBench/Bring reason/Code

\paragraph{VLM Domain Adaptation} % 留着, 引出为什么要做融合
% LLM -> VLM finetune

While supervised fine-tuning (SFT) can adapt VLMs to specific domains such as medicine and mathematics~\cite{llavamed, mathllava}, it remains resource-intensive and time-consuming. Parameter-efficient fine-tuning (PEFT) offers an economical alternative by training only a small subset of parameters, such as adapters, LoRA, or soft prompts, while keeping pretrained weights frozen~\cite{lora, qlora, vladapter}. However, PEFT still requires substantial curated image–text pairs, which can be expensive to obtain or scarce in specialized domains. In contrast, \algo{} (discussed in this paper) offers a promising alternative that efficiently transfers specialized capabilities without requiring large-scale vision data or incurring extra training overhead.

% Vision-Language Models typically comprise three main components: a vision encoder, a modality-alignment projector, and a text decoder. Rather than training these models entirely from scratch, current best practices often use pretrained vision encoders and pretrained text decoders. On the text side, open-source Large Language Models (LLMs), such as Mistral, Qwen, and the LLaMA series, are widely adopted and frequently enhanced through supervised fine-tuning (SFT), yielding diverse expert models. This standard architecture forms the foundation for our proposed method, which specifically explores merging expert LLMs into the textual component of VLMs.

% Supervised Finetuning(SFT) is often used to help VLMs obtain better performance is specific domains such as instruction following, math, legal, medical and coding. %citation needed 
% However, finetuning a VLM is resource intensive and time consuming and sometimes. Moreover, domain vision data not easily be available and maybe scarcero sensitive. Merging VLM backbone and expert LLM may help VLMs obtain abilities are readily.

\paragraph{VLM Merging}

With the growing prominence of VLMs, model merging has been extended from purely language-based settings to multimodal contexts. Most existing work focuses on VLM-to-VLM merging, whether under homogeneous or heterogeneous backbones, exploring techniques such as uncertainty-guided selection, module-level recipes, fine-grained parameter splicing, and cross-architecture alignment \citep{UQ-Merge, remedy, graft, AdaMMS}.

More closely related to our work is the VLM-LLM merging paradigm, where a domain-specialized LLM is merged into a VLM backbone to induce cross-modal capabilities. \citet{bringreasontovision} inject reasoning abilities into VLMs and investigate how perception and reasoning are distributed across layers. \citet{viscodex} integrate a code-specialized LLM with a VLM to enable multimodal code generation. \citet{TransferringTextual} merge text-based reward models into VLMs to construct vision-language reward models that preserve the original preferences.

However, these prior efforts remain task-specific (targeting reasoning, code, or reward modeling) and do not systematically examine when \algo{} succeeds, which merging methods are most effective, or how sensitive performance is to hyperparameter choices. In contrast, we present a comprehensive study across diverse scenarios, merging methods, and hyperparameter choices within a unified framework.

% The most intuitive method of model merging is Simple Weight Averaging (SWA), where the parameters of multiple models are combined linearly. However, this naive approach frequently leads to parameter or task interference, failing to preserve the specialized abilities of individual models.

% Recent studies have proposed advanced methods to mitigate these interference issues. For instance, pruning-based techniques such as TIES-Merging \cite{yadav2023tiesmerging} selectively prune insignificant parameter updates and resolve conflicts in task vectors (changes relative to a base model post-finetuning) \cite{ilharco2023taskarithmetic}. Similarly, DARE \cite{yu2024languagemodelssupermario} employs a randomized dropping and scaling approach to preserve critical task-specific capabilities.

% Data-aware merging strategies have also shown significant promise.  Activation-Informed Merging (AIM)\cite{heyrani2025aim}  employs a small calibration set to identify critical neurons via their activations to preserve important model parameters. Fisher merge computes per-parameter Fisher information weights using a small dataset, resulting in a data-driven weighted parameter averaging approach.

% Additionally, merging at layer granularity has proven effective. Techniques such as LayerSwap \cite{bandarkar2024layerswap} replace specific layers of one LLM with those from another to combine capabilities with minimal interference.

\section{Conclusion}

We systematically investigate cross-modal skill injection for transferring expert LLM capabilities to VLMs. Our experiments reveal that language and instruction-following abilities transfer effectively, while mathematical reasoning remains challenging. Among merging methods, classic merging (i.e., TA, DARE) achieves the best performance, and NaN offers low-cost tuning-free exploration. GP-BO proves most effective for hyperparameter optimization. Our findings provide practical guidelines for efficient cross-modal model merging.

\section*{Limitations}

Due to computational resource constraints, our study focuses exclusively on visual--language modalities and does not extend to other modalities such as audio or video. We leave the investigation of cross-modal skill injection across a broader range of modalities to future work.

\section*{Risks and Ethical Considerations}

This work is primarily methodological and does not involve user data or deployment in high-stakes decision-making. We do not identify significant direct risks. We encourage future work to assess downstream fairness, privacy, and safety impacts in application-specific settings.

\section*{Acknowledgements}

We thank the anonymous reviewers for their insightful comments and helpful suggestions. This work was supported in part by the National Natural Science Foundation of China under Grant No.~92470205. Xu Sun is the corresponding author.

\bibliography{custom}

@inproceedings{lora,
    title={Lo{RA}: Low-Rank Adaptation of Large Language Models},
    author={Edward J Hu and yelong shen and Phillip Wallis and Zeyuan Allen-Zhu and Yuanzhi Li and Shean Wang and Lu Wang and Weizhu Chen},
    booktitle={International Conference on Learning Representations},
    year={2022},
    url={https://openreview.net/forum?id=nZeVKeeFYf9}
}

@inproceedings{qlora,
    title={{QL}o{RA}: Efficient Finetuning of Quantized {LLM}s},
    author={Tim Dettmers and Artidoro Pagnoni and Ari Holtzman and Luke Zettlemoyer},
    booktitle={Thirty-seventh Conference on Neural Information Processing Systems},
    year={2023},
    url={https://openreview.net/forum?id=OUIFPHEgJU}
}

@InProceedings{llava,
    author    = {Liu, Haotian and Li, Chunyuan and Li, Yuheng and Lee, Yong Jae},
    title     = {Improved Baselines with Visual Instruction Tuning},
    booktitle = {Proceedings of the IEEE/CVF Conference on Computer Vision and Pattern Recognition (CVPR)},
    month     = {June},
    year      = {2024},
    pages     = {26296-26306}
}

@inproceedings{BLIP-2,
    author = {Li, Junnan and Li, Dongxu and Savarese, Silvio and Hoi, Steven},
    title = {BLIP-2: bootstrapping language-image pre-training with frozen image encoders and large language models},
    year = {2023},
    publisher = {JMLR.org},
    booktitle = {Proceedings of the 40th International Conference on Machine Learning},
    articleno = {814},
    numpages = {13},
    location = {Honolulu, Hawaii, USA},
    series = {ICML'23}
}

@inproceedings{flamingo,
 author = {Alayrac, Jean-Baptiste and Donahue, Jeff and Luc, Pauline and Miech, Antoine and Barr, Iain and Hasson, Yana and Lenc, Karel and Mensch, Arthur and Millican, Katherine and Reynolds, Malcolm and Ring, Roman and Rutherford, Eliza and Cabi, Serkan and Han, Tengda and Gong, Zhitao and Samangooei, Sina and Monteiro, Marianne and Menick, Jacob L and Borgeaud, Sebastian and Brock, Andy and Nematzadeh, Aida and Sharifzadeh, Sahand and Bi\'{n}kowski, Miko\l aj and Barreira, Ricardo and Vinyals, Oriol and Zisserman, Andrew and Simonyan, Kar\'{e}n},
 booktitle = {Advances in Neural Information Processing Systems},
 editor = {S. Koyejo and S. Mohamed and A. Agarwal and D. Belgrave and K. Cho and A. Oh},
 pages = {23716--23736},
 publisher = {Curran Associates, Inc.},
 title = {Flamingo: a Visual Language Model for Few-Shot Learning},
 url = {https://proceedings.neurips.cc/paper_files/paper/2022/file/960a172bc7fbf0177ccccbb411a7d800-Paper-Conference.pdf},
 volume = {35},
 year = {2022}
}

@inproceedings{mathvista,
title={MathVista: Evaluating Mathematical Reasoning of Foundation Models in Visual Contexts},
author={Pan Lu and Hritik Bansal and Tony Xia and Jiacheng Liu and Chunyuan Li and Hannaneh Hajishirzi and Hao Cheng and Kai-Wei Chang and Michel Galley and Jianfeng Gao},
booktitle={The Twelfth International Conference on Learning Representations},
year={2024},
url={https://openreview.net/forum?id=KUNzEQMWU7}
}

@inproceedings{miabench,
title={{MIA}-Bench: Towards Better Instruction Following Evaluation of Multimodal {LLM}s},
author={Yusu Qian and Hanrong Ye and Jean-Philippe Fauconnier and Peter Grasch and Yinfei Yang and Zhe Gan},
booktitle={The Thirteenth International Conference on Learning Representations},
year={2025},
url={https://openreview.net/forum?id=7EhS3YBxjY}
}

@inproceedings{mathverse,
author = {Zhang, Renrui and Jiang, Dongzhi and Zhang, Yichi and Lin, Haokun and Guo, Ziyu and Qiu, Pengshuo and Zhou, Aojun and Lu, Pan and Chang, Kai-Wei and Qiao, Yu and Gao, Peng and Li, Hongsheng},
title = {MATHVERSE: Does Your Multi-modal LLM Truly See the Diagrams in Visual Math Problems?},
year = {2024},
isbn = {978-3-031-73241-6},
publisher = {Springer-Verlag},
address = {Berlin, Heidelberg},
url = {https://doi.org/10.1007/978-3-031-73242-3_10},
doi = {10.1007/978-3-031-73242-3_10},
booktitle = {Computer Vision – ECCV 2024: 18th European Conference, Milan, Italy, September 29–October 4, 2024, Proceedings, Part VIII},
pages = {169–186},
numpages = {18},
location = {Milan, Italy}
}

@inproceedings{jmmmu,
    title = "{JMMMU}: A {J}apanese Massive Multi-discipline Multimodal Understanding Benchmark for Culture-aware Evaluation",
    author = "Onohara, Shota  and
      Miyai, Atsuyuki  and
      Imajuku, Yuki  and
      Egashira, Kazuki  and
      Baek, Jeonghun  and
      Yue, Xiang  and
      Neubig, Graham  and
      Aizawa, Kiyoharu",
    editor = "Chiruzzo, Luis  and
      Ritter, Alan  and
      Wang, Lu",
    booktitle = "Proceedings of the 2025 Conference of the Nations of the Americas Chapter of the Association for Computational Linguistics: Human Language Technologies (Volume 1: Long Papers)",
    month = apr,
    year = "2025",
    address = "Albuquerque, New Mexico",
    publisher = "Association for Computational Linguistics",
    url = "https://aclanthology.org/2025.naacl-long.43/",
    doi = "10.18653/v1/2025.naacl-long.43",
    pages = "932--950",
    ISBN = "979-8-89176-189-6"
}

@misc{cmmmu,
      title={CMMMU: A Chinese Massive Multi-discipline Multimodal Understanding Benchmark}, 
      author={Ge Zhang and Xinrun Du and Bei Chen and Yiming Liang and Tongxu Luo and Tianyu Zheng and Kang Zhu and Yuyang Cheng and Chunpu Xu and Shuyue Guo and Haoran Zhang and Xingwei Qu and Junjie Wang and Ruibin Yuan and Yizhi Li and Zekun Wang and Yudong Liu and Yu-Hsuan Tsai and Fengji Zhang and Chenghua Lin and Wenhao Huang and Jie Fu},
      year={2024},
      eprint={2401.11944},
      archivePrefix={arXiv},
      primaryClass={cs.CL},
      url={https://arxiv.org/abs/2401.11944}, 
}

@misc{wildvision,
      title={WildVision: Evaluating Vision-Language Models in the Wild with Human Preferences}, 
      author={Yujie Lu and Dongfu Jiang and Wenhu Chen and William Yang Wang and Yejin Choi and Bill Yuchen Lin},
      year={2024},
      eprint={2406.11069},
      archivePrefix={arXiv},
      primaryClass={cs.CV},
      url={https://arxiv.org/abs/2406.11069}, 
}

@inproceedings{fisher,
author = {Matena, Michael and Raffel, Colin},
title = {Merging models with fisher-weighted averaging},
year = {2022},
isbn = {9781713871088},
publisher = {Curran Associates Inc.},
address = {Red Hook, NY, USA},
booktitle = {Proceedings of the 36th International Conference on Neural Information Processing Systems},
articleno = {1287},
numpages = {14},
location = {New Orleans, LA, USA},
series = {NIPS '22}
}

@inproceedings{regmean,
title={Dataless Knowledge Fusion by Merging Weights of Language Models},
author={Xisen Jin and Xiang Ren and Daniel Preotiuc-Pietro and Pengxiang Cheng},
booktitle={The Eleventh International Conference on Learning Representations },
year={2023},
url={https://openreview.net/forum?id=FCnohuR6AnM}
}

@inproceedings{dare,
author = {Yu, Le and Yu, Bowen and Yu, Haiyang and Huang, Fei and Li, Yongbin},
title = {Language models are super mario: absorbing abilities from homologous models as a free lunch},
year = {2024},
publisher = {JMLR.org},
booktitle = {Proceedings of the 41st International Conference on Machine Learning},
articleno = {2382},
numpages = {21},
location = {Vienna, Austria},
series = {ICML'24}
}

@inproceedings{ties,
title={{TIES}-Merging: Resolving Interference When Merging Models},
author={Prateek Yadav and Derek Tam and Leshem Choshen and Colin Raffel and Mohit Bansal},
booktitle={Thirty-seventh Conference on Neural Information Processing Systems},
year={2023},
url={https://openreview.net/forum?id=xtaX3WyCj1}
}

@inproceedings{ta,
title={Editing models with task arithmetic},
author={Gabriel Ilharco and Marco Tulio Ribeiro and Mitchell Wortsman and Ludwig Schmidt and Hannaneh Hajishirzi and Ali Farhadi},
booktitle={The Eleventh International Conference on Learning Representations },
year={2023},
url={https://openreview.net/forum?id=6t0Kwf8-jrj}
}

@inproceedings{metagpt,
    title = "{M}eta{GPT}: Merging Large Language Models Using Model Exclusive Task Arithmetic",
    author = "Zhou, Yuyan  and
      Song, Liang  and
      Wang, Bingning  and
      Chen, Weipeng",
    editor = "Al-Onaizan, Yaser  and
      Bansal, Mohit  and
      Chen, Yun-Nung",
    booktitle = "Proceedings of the 2024 Conference on Empirical Methods in Natural Language Processing",
    month = nov,
    year = "2024",
    address = "Miami, Florida, USA",
    publisher = "Association for Computational Linguistics",
    url = "https://aclanthology.org/2024.emnlp-main.102/",
    doi = "10.18653/v1/2024.emnlp-main.102",
    pages = "1711--1724"
}

@misc{nan,
      title={NAN: A Training-Free Solution to Coefficient Estimation in Model Merging}, 
      author={Chongjie Si and Kangtao Lv and Jingjing Jiang and Yadao Wang and Yongwei Wang and Xiaokang Yang and Wenbo Su and Bo Zheng and Wei Shen},
      year={2025},
      eprint={2505.16148},
      archivePrefix={arXiv},
      primaryClass={cs.LG},
      url={https://arxiv.org/abs/2505.16148}, 
}

@inproceedings{wudi,
title={Whoever Started the interference Should End It: Guiding Data-Free Model Merging via Task Vectors},
author={Runxi Cheng and Feng Xiong and Yongxian Wei and Wanyun Zhu and Chun Yuan},
booktitle={Forty-second International Conference on Machine Learning},
year={2025},
url={https://openreview.net/forum?id=xR9msNaREW}
}

@article{tsv,
  title={Task Singular Vectors: Reducing Task Interference in Model Merging},
  author={Antonio Andrea Gargiulo and Donato Crisostomi and Maria Sofia Bucarelli and Simone Scardapane and Fabrizio Silvestri and Emanuele Rodol{\`a}},
  journal={2025 IEEE/CVF Conference on Computer Vision and Pattern Recognition (CVPR)},
  year={2024},
  pages={18695-18705},
  url={https://api.semanticscholar.org/CorpusID:274436302}
}

@inproceedings{breadcrumbs,
title={Model Breadcrumbs: Scalable Upcycling of Finetuned Foundation Models via Sparse Task Vectors Merging},
author={MohammadReza Davari and Eugene Belilovsky},
booktitle={ICML 2024 Workshop on Foundation Models in the Wild},
year={2024},
url={https://openreview.net/forum?id=vuyP3tupig}
}

@misc{della,
      title={DELLA-Merging: Reducing Interference in Model Merging through Magnitude-Based Sampling}, 
      author={Pala Tej Deep and Rishabh Bhardwaj and Soujanya Poria},
      year={2024},
      eprint={2406.11617},
      archivePrefix={arXiv},
      primaryClass={cs.CL},
      url={https://arxiv.org/abs/2406.11617}, 
}

@inproceedings{mergekit,
    title = "Arcee{'}s {M}erge{K}it: A Toolkit for Merging Large Language Models",
    author = "Goddard, Charles  and
      Siriwardhana, Shamane  and
      Ehghaghi, Malikeh  and
      Meyers, Luke  and
      Karpukhin, Vladimir  and
      Benedict, Brian  and
      McQuade, Mark  and
      Solawetz, Jacob",
    editor = "Dernoncourt, Franck  and
      Preo{\c{t}}iuc-Pietro, Daniel  and
      Shimorina, Anastasia",
    booktitle = "Proceedings of the 2024 Conference on Empirical Methods in Natural Language Processing: Industry Track",
    month = nov,
    year = "2024",
    address = "Miami, Florida, US",
    publisher = "Association for Computational Linguistics",
    url = "https://aclanthology.org/2024.emnlp-industry.36/",
    doi = "10.18653/v1/2024.emnlp-industry.36",
    pages = "477--485"
}

@InProceedings{bringreasontovision,
  title = 	 {Bring Reason to Vision: Understanding Perception and Reasoning through Model Merging},
  author =       {Chen, Shiqi and Zhang, Jinghan and Zhu, Tongyao and Liu, Wei and Gao, Siyang and Xiong, Miao and Li, Manling and He, Junxian},
  booktitle = 	 {Proceedings of the 42nd International Conference on Machine Learning},
  pages = 	 {9803--9817},
  year = 	 {2025},
  editor = 	 {Singh, Aarti and Fazel, Maryam and Hsu, Daniel and Lacoste-Julien, Simon and Berkenkamp, Felix and Maharaj, Tegan and Wagstaff, Kiri and Zhu, Jerry},
  volume = 	 {267},
  series = 	 {Proceedings of Machine Learning Research},
  month = 	 {13--19 Jul},
  publisher =    {PMLR},
  url = 	 {https://proceedings.mlr.press/v267/chen25cm.html}
}

@misc{lmmsevalrealitycheckevaluation,
      title={LMMs-Eval: Reality Check on the Evaluation of Large Multimodal Models}, 
      author={Kaichen Zhang and Bo Li and Peiyuan Zhang and Fanyi Pu and Joshua Adrian Cahyono and Kairui Hu and Shuai Liu and Yuanhan Zhang and Jingkang Yang and Chunyuan Li and Ziwei Liu},
      year={2024},
      eprint={2407.12772},
      archivePrefix={arXiv},
      primaryClass={cs.CL},
      url={https://arxiv.org/abs/2407.12772}, 
}

@misc{lmmseval,
    title={LMMs-Eval: Accelerating the Development of Large Multimoal Models},
    url={https://github.com/EvolvingLMMs-Lab/lmms-eval},
    author={Bo Li and Peiyuan Zhang and Kaichen Zhang and Fanyi Pu and Xinrun Du and Yuhao Dong and Haotian Liu and Yuanhan Zhang and Ge Zhang and Chunyuan Li and Ziwei Liu},
    publisher    = {Zenodo},
    version      = {v0.1.0},
    month={March},
    year={2024}
}

@misc{mistral,
      title={Mistral 7B}, 
      author={Albert Q. Jiang and Alexandre Sablayrolles and Arthur Mensch and Chris Bamford and Devendra Singh Chaplot and Diego de las Casas and Florian Bressand and Gianna Lengyel and Guillaume Lample and Lucile Saulnier and Lélio Renard Lavaud and Marie-Anne Lachaux and Pierre Stock and Teven Le Scao and Thibaut Lavril and Thomas Wang and Timothée Lacroix and William El Sayed},
      year={2023},
      eprint={2310.06825},
      archivePrefix={arXiv},
      primaryClass={cs.CL},
      url={https://arxiv.org/abs/2310.06825}, 
}

@inproceedings{datamix,
title={Data Mixing Optimization for Supervised Fine-Tuning of Large Language Models},
author={Yuan Li and Zhengzhong Liu and Eric P. Xing},
booktitle={Forty-second International Conference on Machine Learning},
year={2025},
url={https://openreview.net/forum?id=19kqoNoc2N}
}

@inproceedings{mammothvl,
    title = "{MA}mmo{TH}-{VL}: Eliciting Multimodal Reasoning with Instruction Tuning at Scale",
    author = "Guo, Jiawei  and
      Zheng, Tianyu  and
      Li, Yizhi  and
      Bai, Yuelin  and
      Li, Bo  and
      Wang, Yubo  and
      Zhu, King  and
      Neubig, Graham  and
      Chen, Wenhu  and
      Yue, Xiang",
    editor = "Che, Wanxiang  and
      Nabende, Joyce  and
      Shutova, Ekaterina  and
      Pilehvar, Mohammad Taher",
    booktitle = "Proceedings of the 63rd Annual Meeting of the Association for Computational Linguistics (Volume 1: Long Papers)",
    month = jul,
    year = "2025",
    address = "Vienna, Austria",
    publisher = "Association for Computational Linguistics",
    url = "https://aclanthology.org/2025.acl-long.680/",
    doi = "10.18653/v1/2025.acl-long.680",
    pages = "13869--13920",
    ISBN = "979-8-89176-251-0"
}

@inproceedings{wit,
author = {Srinivasan, Krishna and Raman, Karthik and Chen, Jiecao and Bendersky, Michael and Najork, Marc},
title = {WIT: Wikipedia-based Image Text Dataset for Multimodal Multilingual Machine Learning},
year = {2021},
isbn = {9781450380379},
publisher = {Association for Computing Machinery},
address = {New York, NY, USA},
url = {https://doi.org/10.1145/3404835.3463257},
doi = {10.1145/3404835.3463257},
booktitle = {Proceedings of the 44th International ACM SIGIR Conference on Research and Development in Information Retrieval},
pages = {2443–2449},
numpages = {7},
location = {Virtual Event, Canada},
series = {SIGIR '21}
}

@inproceedings{scenario_work1,
    title = "Extracting and Combining Abilities For Building Multi-lingual Ability-enhanced Large Language Models",
    author = "Chen, Zhipeng  and
      Zhou, Kun  and
      Song, Liang  and
      Zhao, Wayne Xin  and
      Wang, Bingning  and
      Chen, Weipeng  and
      Wen, Ji-Rong",
    editor = "Christodoulopoulos, Christos  and
      Chakraborty, Tanmoy  and
      Rose, Carolyn  and
      Peng, Violet",
    booktitle = "Proceedings of the 2025 Conference on Empirical Methods in Natural Language Processing",
    month = nov,
    year = "2025",
    address = "Suzhou, China",
    publisher = "Association for Computational Linguistics",
    url = "https://aclanthology.org/2025.emnlp-main.887/",
    doi = "10.18653/v1/2025.emnlp-main.887",
    pages = "17574--17591",
    ISBN = "979-8-89176-332-6"
}

@article{scenario_work2,
  title={RCP-Merging: Merging Long Chain-of-Thought Models with Domain-Specific Models by Considering Reasoning Capability as Prior},
  author={Junyao Yang and Jianwei Wang and Huiping Zhuang and Cen Chen and Ziqian Zeng},
  journal={ArXiv},
  year={2025},
  volume={abs/2508.03140},
  url={https://api.semanticscholar.org/CorpusID:280526616}
}

@inproceedings{idefics2,
  title={What matters when building vision-language models?},
  author={Hugo Laurencon and Leo Tronchon and Matthieu Cord and Victor Sanh},
  booktitle={The Thirty-eighth Annual Conference on Neural Information Processing Systems},
  year={2024},
  url={https://openreview.net/forum?id=dtvJF1Vy2i}
}

@inproceedings{llava_model,
  title={Visual Instruction Tuning},
  author={Haotian Liu and Chunyuan Li and Qingyang Wu and Yong Jae Lee},
  booktitle={Thirty-seventh Conference on Neural Information Processing Systems},
  year={2023},
  url={https://openreview.net/forum?id=w0H2xGHlkw}
}

@article{qwen2vl,
  title={Qwen2-VL: Enhancing Vision-Language Model's Perception of the World at Any Resolution},
  author={Peng Wang and Shuai Bai and Sinan Tan and Shijie Wang and Zhihao Fan and Jinze Bai and Ke-Yang Chen and Xuejing Liu and Jialin Wang and Wenbin Ge and Yang Fan and Kai Dang and Mengfei Du and Xuancheng Ren and Rui Men and Dayiheng Liu and Chang Zhou and Jingren Zhou and Junyang Lin},
  journal={ArXiv},
  year={2024},
  volume={abs/2409.12191},
  url={https://api.semanticscholar.org/CorpusID:272704132}
}

@misc{alpaca_zh,
      title={Efficient and Effective Text Encoding for Chinese LLaMA and Alpaca}, 
      author={Yiming Cui and Ziqing Yang and Xin Yao},
      year={2024},
      eprint={2304.08177},
      archivePrefix={arXiv},
      primaryClass={cs.CL},
      url={https://arxiv.org/abs/2304.08177}, 
}

@misc{alpaca_ja,
  title        = {japanese\_alpaca\_data},
  author       = {{fujiki}},
  howpublished = {Hugging Face Datasets},
  url          = {https://huggingface.co/datasets/fujiki/japanese_alpaca_data},
  year={2023},
  note         = {Accessed 2026-01-01}
}

@inproceedings{math500,
title={Let's Verify Step by Step},
author={Hunter Lightman and Vineet Kosaraju and Yuri Burda and Harrison Edwards and Bowen Baker and Teddy Lee and Jan Leike and John Schulman and Ilya Sutskever and Karl Cobbe},
booktitle={The Twelfth International Conference on Learning Representations},
year={2024},
url={https://openreview.net/forum?id=v8L0pN6EOi}
}

@misc{dolly,
    author    = {Mike Conover and Matt Hayes and Ankit Mathur and Jianwei Xie and Jun Wan and Sam Shah and Ali Ghodsi and Patrick Wendell and Matei Zaharia and Reynold Xin},
    title     = {Free Dolly: Introducing the World's First Truly Open Instruction-Tuned LLM},
    year      = {2023},
    url       = {https://www.databricks.com/blog/2023/04/12/dolly-first-open-commercially-viable-instruction-tuned-llm},
    urldate   = {2023-06-30}
}

@misc{llavanext,
    title={LLaVA-NeXT: Improved reasoning, OCR, and world knowledge},
    url={https://llava-vl.github.io/blog/2024-01-30-llava-next/},
    author={Liu, Haotian and Li, Chunyuan and Li, Yuheng and Li, Bo and Zhang, Yuanhan and Shen, Sheng and Lee, Yong Jae},
    month={January},
    year={2024}
}

@misc{improvedllava,
      title={Improved Baselines with Visual Instruction Tuning}, 
      author={Liu, Haotian and Li, Chunyuan and Li, Yuheng and Lee, Yong Jae},
      publisher={arXiv:2310.03744},
      year={2023},
}

@inproceedings{UQ-Merge,
    title = "UQ-Merge: Uncertainty Guided Multimodal Large Language Model Merging",
    author = "Qu, Huaizhi  and
      Zhao, Xinyu  and
      Peng, Jie  and
      Lee, Kwonjoon  and
      Dariush, Behzad  and
      Chen, Tianlong",
    editor = "Che, Wanxiang  and
      Nabende, Joyce  and
      Shutova, Ekaterina  and
      Pilehvar, Mohammad Taher",
    booktitle = "Findings of the Association for Computational Linguistics: ACL 2025",
    month = jul,
    year = "2025",
    address = "Vienna, Austria",
    publisher = "Association for Computational Linguistics",
    url = "https://aclanthology.org/2025.findings-acl.73/",
    doi = "10.18653/v1/2025.findings-acl.73",
    pages = "1401--1417",
    ISBN = "979-8-89176-256-5"
}

@inproceedings{remedy,
title={{REMEDY}: Recipe Merging Dynamics in Large Vision-Language Models},
author={Didi Zhu and Yibing Song and Tao Shen and Ziyu Zhao and Jinluan Yang and Min Zhang and Chao Wu},
booktitle={The Thirteenth International Conference on Learning Representations},
year={2025},
url={https://openreview.net/forum?id=iX7eHHE5Tx}
}

@article{graft,
  title={Graft: Integrating the Domain Knowledge via Efficient Parameter Synergy for MLLMs},
  author={Yang Dai and Jianxiang An and Tianwei Lin and Hongyang He and Hongzhe Huang and Wenqiao Zhang and Zheqi Lv and Siliang Tang and Yueting Zhuang},
  journal={ArXiv},
  year={2025},
  volume={abs/2506.23940},
  url={https://api.semanticscholar.org/CorpusID:280011454}
}

@article{AdaMMS,
  publtype={informal},
  author={Yiyang Du and Xiaochen Wang and Chi Chen and Jiabo Ye and Yiru Wang and Peng Li and Ming Yan and Ji Zhang and Fei Huang and Zhifang Sui and Maosong Sun and Yang Liu},
  title={AdaMMS: Model Merging for Heterogeneous Multimodal Large Language Models with Unsupervised Coefficient Optimization},
  year={2025},
  month={March},
  cdate={1740787200000},
  journal={CoRR},
  volume={abs/2503.23733},
  url={https://doi.org/10.48550/arXiv.2503.23733}
}

@inproceedings{viscodex,
title={VisCodex: Unified Multimodal Code Generation via Merging Vision and Coding Models},
author={Lingjie Jiang and Shaohan Huang and Xun Wu and Yixia Li and Guanhua Chen and Dongdong Zhang and Furu Wei},
booktitle={The Fourteenth International Conference on Learning Representations},
year={2026},
url={https://openreview.net/forum?id=hUXzPauNEM}
}

@inproceedings{TransferringTextual,
    title = "Transferring Textual Preferences to Vision-Language Understanding through Model Merging",
    author = "Li, Chen-An  and
      Lin, Tzu-Han  and
      Chen, Yun-Nung  and
      Lee, Hung-yi",
    editor = "Che, Wanxiang  and
      Nabende, Joyce  and
      Shutova, Ekaterina  and
      Pilehvar, Mohammad Taher",
    booktitle = "Proceedings of the 63rd Annual Meeting of the Association for Computational Linguistics (Volume 2: Short Papers)",
    month = jul,
    year = "2025",
    address = "Vienna, Austria",
    publisher = "Association for Computational Linguistics",
    url = "https://aclanthology.org/2025.acl-short.72/",
    doi = "10.18653/v1/2025.acl-short.72",
    pages = "923--943",
    ISBN = "979-8-89176-252-7"
}

@article{directsearch,
  title={Optimization by direct search: New perspectives on some classical and modern methods},
  author={Kolda, Tamara G and Lewis, Robert Michael and Torczon, Virginia},
  journal={SIAM review},
  volume={45},
  number={3},
  pages={385--482},
  year={2003},
  publisher={SIAM}
}

@article{sobol,
title = {On the distribution of points in a cube and the approximate evaluation of integrals},
journal = {USSR Computational Mathematics and Mathematical Physics},
volume = {7},
number = {4},
pages = {86-112},
year = {1967},
issn = {0041-5553},
doi = {https://doi.org/10.1016/0041-5553(67)90144-9},
url = {https://www.sciencedirect.com/science/article/pii/0041555367901449},
author = {I.M Sobol}
}

@article{cmaes,
  title={Completely Derandomized Self-Adaptation in Evolution Strategies},
  author={Nikolaus Hansen and Andreas Ostermeier},
  journal={Evolutionary Computation},
  year={2001},
  volume={9},
  pages={159-195},
  url={https://api.semanticscholar.org/CorpusID:7524826}
}

@article{gpbo,
  title={Efficient global optimization of expensive black-box functions},
  author={Jones, Donald R and Schonlau, Matthias and Welch, William J},
  journal={Journal of Global optimization},
  volume={13},
  number={4},
  pages={455--492},
  year={1998},
  publisher={Springer}
}

@article{pattern,
  title={On the convergence of pattern search algorithms},
  author={Torczon, Virginia},
  journal={SIAM Journal on optimization},
  volume={7},
  number={1},
  pages={1--25},
  year={1997},
  publisher={SIAM}
}

@article{powell,
  title={An efficient method for finding the minimum of a function of several variables without calculating derivatives},
  author={Powell, Michael JD},
  journal={The computer journal},
  volume={7},
  number={2},
  pages={155--162},
  year={1964},
  publisher={Oxford University Press}
}

@inproceedings{llavamed,
title={LLaVA-Med: Training a Large Language-and-Vision Assistant for Biomedicine in One Day},
author={Chunyuan Li and Cliff Wong and Sheng Zhang and Naoto Usuyama and Haotian Liu and Jianwei Yang and Tristan Naumann and Hoifung Poon and Jianfeng Gao},
booktitle={Thirty-seventh Conference on Neural Information Processing Systems Datasets and Benchmarks Track},
year={2023},
url={https://openreview.net/forum?id=GSuP99u2kR}
}

@inproceedings{mathllava,
    title = "Math-LLaVA: Bootstrapping Mathematical Reasoning for Multimodal Large Language Models",
    author = "Shi, Wenhao  and
      Hu, Zhiqiang  and
      Bin, Yi  and
      Liu, Junhua  and
      Yang, Yang  and
      Ng, See-Kiong  and
      Bing, Lidong  and
      Lee, Roy Ka-Wei",
    editor = "Al-Onaizan, Yaser  and
      Bansal, Mohit  and
      Chen, Yun-Nung",
    booktitle = "Findings of the Association for Computational Linguistics: EMNLP 2024",
    month = nov,
    year = "2024",
    address = "Miami, Florida, USA",
    publisher = "Association for Computational Linguistics",
    url = "https://aclanthology.org/2024.findings-emnlp.268/",
    doi = "10.18653/v1/2024.findings-emnlp.268",
    pages = "4663--4680"
}

@InProceedings{vladapter,
    author    = {Sung, Yi-Lin and Cho, Jaemin and Bansal, Mohit},
    title     = {VL-Adapter: Parameter-Efficient Transfer Learning for Vision-and-Language Tasks},
    booktitle = {Proceedings of the IEEE/CVF Conference on Computer Vision and Pattern Recognition (CVPR)},
    month     = {June},
    year      = {2022},
    pages     = {5227-5237}
}

@article{gsm8k,
  title={Training Verifiers to Solve Math Word Problems},
  author={Cobbe, Karl and Kosaraju, Vineet and Bavarian, Mohammad and Chen, Mark and Jun, Heewoo and Kaiser, Lukasz and Plappert, Matthias and Tworek, Jerry and Hilton, Jacob and Nakano, Reiichiro and Hesse, Christopher and Schulman, John},
  journal={arXiv preprint arXiv:2110.14168},
  year={2021}
}

\clearpage

\appendix
\section*{Appendix}
\addcontentsline{toc}{section}{Appendix}

\section{The models used in experiments}
\label{sec:checkpoint}

Table~\ref{tab:appendix_models} summarizes the models used in our experiments across different scenarios. For each scenario, we pair a VLM with a domain-expert LLM that shares the same base architecture. Specifically, for language understanding tasks, we select Chinese (Mistral-7B-v0.3-Chinese-Chat) and Japanese (Llama-3-ELYZA-JP-8B). For mathematical reasoning, we use the DART-Math series, which are LLaMA and Mistral models fine-tuned on mathematical problem-solving data. For instruction following, we pair Qwen2-VL with Qwen2-Instruct, and Idefics2-base with Mistral-7B-Instruct. All models are publicly available on Hugging Face.

\section{Hyperparameters Used}
\label{sec:hyper}

Table~\ref{tab:algo_hparams} details the hyperparameter configurations in Section~\ref{sec::merging} for each merging method. Classic merging methods (Task Arithmetic, DARE, TIES) require tuning merging coefficients for both the VLM and expert LLM components. We perform grid search over coefficient values in $\{0.1, 0.3, 0.5, 0.7, 0.9\}$. For DARE and TIES, which incorporate sparsification, we additionally search over density values in $\{0.2, 0.4, 0.6, 0.8\}$. This results in 25 configurations for Task Arithmetic and 100 configurations for DARE and TIES.

Data-aware methods (Fisher and RegMean) and tuning-free methods (WUDI, TSV, MetaGPT, NaN) do not require hyperparameter search, as they either derive merging weights from data statistics or compute them directly from model parameters.

\section{Calibration Data for Data-Aware Methods}
\label{sec:data}

For data-aware merging methods (Fisher and RegMean), we use 500 calibration samples randomly drawn from the corresponding dataset (i.e., for merging between base VLMs and math expert LLMs, we calibrate the VLM on a vision dataset and the math expert LLM on a textual mathematical reasoning dataset, respectively). The specific data sources for each scenario are as follows (all datasets can be downloaded from HuggingFace):

\begin{itemize}[leftmargin=1em]
    \item \textbf{Vision Perception:} \texttt{liuhaotian/LLaVA-Ins\-truct-150K}, a large-scale visual instruction tuning dataset.~\citep{llava_model, improvedllava, llavanext}
    \item \textbf{Language (Chinese):} \texttt{hfl/alpaca\_zh\_51k}, a Chinese instruction-following dataset translated and adapted from the original Alpaca dataset.~\citep{alpaca_zh}
    \item \textbf{Language (Japanese):} \texttt{fujiki/japanese\_al\-paca\_data}, a Japanese version of the Alpaca dataset for instruction tuning.~\citep{alpaca_ja}
    \item \textbf{Mathematical Reasoning:} \texttt{qwedsacf/compe\-tition\_math}, a subset of challenging mathematical problems requiring multi-step reasoning, combined with \texttt{openai/gsm8k}. We mix the two datasets in a 50/50 ratio and shuffle the samples.~\citep{math500, gsm8k}
    \item \textbf{Instruction Following:} \texttt{databricks/databri\-cks-dolly-15k}, a high-quality instruction-following dataset created by Databricks.~\citep{dolly}
\end{itemize}

\begin{table*}[htbp]
\centering
\small
\setlength{\tabcolsep}{3pt}
\renewcommand{\arraystretch}{1.2}

\begin{tabularx}{\textwidth}{
  >{\raggedright\arraybackslash}l
  >{\centering\arraybackslash}X
  >{\centering\arraybackslash}X
  >{\centering\arraybackslash}X
}
\toprule
\textbf{Scenario} & \textbf{Expert LLM} & \textbf{VLM} \\
\midrule

\multirow{2}{*}{Language} 
    & shenzhi-wang/Mistral-7B-v0.3-Chinese-Chat & llava-hf/llava-v1.6-mistral-7b-hf \\
    & elyza/Llama-3-ELYZA-JP-8B & lmms-lab/llama3-llava-next-8b \\

\midrule

\multirow{2}{*}{Math} 
    & hkust-nlp/dart-math-llama3-8b-prop2diff & lmms-lab/llama3-llava-next-8b \\
    & hkust-nlp/dart-math-mistral-7b-prop2diff & llava-hf/llava-v1.6-mistral-7b-hf \\

\midrule

\multirow{2}{*}{Instruction Following} 
    & Qwen/Qwen2-7B-Instruct & Qwen/Qwen2-VL-7B \\
    & mistralai/Mistral-7B-Instruct-v0.1 & 
 HuggingFaceM4/idefics2-8b-base \\

\bottomrule
\end{tabularx}
\caption{Expert LLMs and VLMs used in our experiments across different scenarios. Each VLM is paired with a domain-expert LLM that shares the same base architecture.}
\label{tab:appendix_models}
\end{table*}

\begin{table*}[htbp]
\centering
\small
\renewcommand{\arraystretch}{1.25}

\begin{tabularx}{\textwidth}{
  p{0.26\textwidth}
  >{\raggedright\arraybackslash}p{0.20\textwidth}
  >{\raggedright\arraybackslash}X
}
\toprule
\textbf{Category} & \textbf{Algorithm} & \textbf{Hyperparameters} \\
\midrule
\multirow{8}{*}{Classic merging}
  & Task Arithmetic
  & \texttt{vl\_weight} $\in \{0.1,0.3,0.5,0.7,0.9\}$ \\[-2pt]
  & 
  & \texttt{exp\_weight} $\in \{0.1,0.3,0.5,0.7,0.9\}$ \\
\cmidrule(l){2-3}
  & DARE
  & \texttt{vl\_weight} $\in \{0.1,0.3,0.5,0.7,0.9\}$ \\[-2pt]
  & 
  & \texttt{exp\_weight} $\in \{0.1,0.3,0.5,0.7,0.9\}$ \\[-2pt]
  & 
  & \texttt{density} $\in \{0.2,0.4,0.6,0.8\}$ \\
\cmidrule(l){2-3}
  & TIES
  & \texttt{vl\_weight} $\in \{0.1,0.3,0.5,0.7,0.9\}$ \\[-2pt]
  & 
  & \texttt{exp\_weight} $\in \{0.1,0.3,0.5,0.7,0.9\}$ \\[-2pt]
  & 
  & \texttt{density} $\in \{0.2,0.4,0.6,0.8\}$ \\
\midrule
\multirow{2}{*}{Data-aware}
  & Fisher Merging & n/a \\
\cmidrule(l){2-3}
  & RegMean        & n/a \\
\midrule
\multirow{4}{*}{Tuning-free}
  & WUDI    & n/a \\
\cmidrule(l){2-3}
  & TSV     & n/a \\
\cmidrule(l){2-3}
  & MetaGPT & n/a \\
\cmidrule(l){2-3}
  & NaN     & n/a \\
\bottomrule
\end{tabularx}

\caption{Overview of merging algorithms and their hyperparameter tuning spaces. Classic merging methods require tuning merging coefficients, while data-aware and tuning-free methods operate without hyperparameter tuning.}
\label{tab:algo_hparams}
\end{table*}

\section{Scaling Results Across Model Sizes}
\label{sec:scaling}

Our main experiments focus on 7B--8B models. This choice is primarily motivated by the current availability of compatible VLM--LLM pairs for cross-modal skill injection: the expert LLM and the VLM backbone must share the same base architecture, and publicly available pairs satisfying this constraint are predominantly concentrated in this size range.

To examine whether our findings generalize beyond this scale, we conduct additional experiments on the Qwen2 family at 2B, 7B, and 72B scales. We evaluate NaN, our recommended tuning-free method, on instruction following, and Task Arithmetic (TA), one of our recommended classic methods, on mathematical reasoning. These results support two observations across model sizes: (1) cross-modal skill injection consistently improves over the corresponding base VLM, and (2) the gains on mathematical reasoning remain markedly smaller than in other scenarios.

Table~\ref{tab:scaling_mia} shows that for instruction following on MIA-Bench, cross-modal skill injection yields substantial gains at every scale. The merged models improve over the base VLM by 29.6 points at 2B, 40.8 points at 7B, and 36.0 points at 72B.

Table~\ref{tab:scaling_math} shows that for mathematical reasoning on the MathVista math subset, improvements also persist across all three scales, but remain much smaller: 0.46 points at 2B, 2.53 points at 7B, and 0.92 points at 72B. Taken together, these results show that the two core patterns identified in our main experiments remain stable across model sizes: merging is beneficial across scales, while mathematical reasoning is consistently harder to transfer than instruction-following ability.

\begin{table}[htbp]
\centering
\small
\begin{tabularx}{\columnwidth}{Xccc}
\toprule
\textbf{Scale} & \textbf{Base VLM} & \textbf{Merged (NaN)} & \textbf{Gain} \\
\midrule
2B  & 29.3 & 58.9 & +29.6 \\
7B  & 33.0 & 73.8 & +40.8 \\
72B & 55.5 & 91.5 & +36.0 \\
\bottomrule
\end{tabularx}
\caption{Scaling results on MIA-Bench. Cross-modal skill injection with NaN improves instruction-following performance across Qwen2 model sizes.}
\label{tab:scaling_mia}
\end{table}

\begin{table}[htbp]
\centering
\small
\begin{tabularx}{\columnwidth}{Xccc}
\toprule
\textbf{Scale} & \textbf{Base VLM} & \textbf{Merged (TA)} & \textbf{Gain} \\
\midrule
2B  & 36.32 & 36.78 & +0.46 \\
7B  & 50.80 & 53.33 & +2.53 \\
72B & 60.46 & 61.38 & +0.92 \\
\bottomrule
\end{tabularx}
\caption{Scaling results on the MathVista math subset. Cross-modal skill injection with Task Arithmetic improves mathematical reasoning across Qwen2 model sizes, but the gains are modest.}
\label{tab:scaling_math}
\end{table}

\section{Full Results on MathVista}
\label{sec:mathvista}

Table~\ref{tab:mathvista_full} presents the complete results on the MathVista benchmark, including the full benchmark as well as its General and Math subsets. On the Math subset, several merging methods achieve small gains over the base VLM, indicating that some mathematical reasoning ability can indeed be injected. For example, RegMean reaches the highest Math-subset score among merged models (27.13 versus 25.29 for the base VLM). However, these gains do not reliably translate to the full benchmark: most merged models underperform the base VLM overall, and even the best merged result remains only comparable.

This gap between the Math subset and the full benchmark provides a plausible explanation for why mathematical transfer remains difficult. The General subset depends more heavily on visual perception and fine-grained visual discrimination, whereas the Math subset more directly rewards symbolic and multi-step reasoning. Merging a math-specialized LLM therefore appears to strengthen reasoning-heavy behavior while simultaneously perturbing perception-related capabilities in the VLM. This interpretation is consistent with the main-text observation that mathematical reasoning can transfer, but the gains are often offset by degraded visual grounding, making mathematical reasoning more challenging.

\begin{table}[htbp]
\centering
\small
\begin{tabularx}{\columnwidth}{Xccc}
\toprule
\textbf{Method} & \textbf{Overall} & \textbf{General} & \textbf{Math} \\
\midrule
TA       & 34.75 & 44.38 & 26.67 \\
DARE     & \textbf{36.12} & \textbf{48.22} & 25.98 \\
TIES     & 32.25 & 41.10 & 24.83 \\
\midrule
Fisher   & 29.88 & 38.90 & 22.30 \\
RegMean  & 30.88 & 35.34 & \textbf{27.13} \\
\midrule
WUDI     & 30.38 & 36.99 & 24.83 \\
MetaGPT  & 20.88 & 23.84 & 18.39 \\
NaN      & 29.62 & 36.71 & 23.68 \\
\midrule
Base VLM & 36.00 & 48.77 & 25.29 \\
\bottomrule
\end{tabularx}
\caption{Performance on MathVista across subsets. Best results among merging methods are in \textbf{bold}.}
\label{tab:mathvista_full}
\end{table}

 \section{Qualitative Examples}
 \label{sec:qualitative_examples}

Here we provide several qualitative examples to help readers better understand the model's capabilities before and after merging.

\textbf{Instruction following:} Merging substantially improves instruction-following performance. In the MIA-Bench example shown in Table~\ref{tab:qualitative_party}, the model is given an image of a party room and asked to describe the atmosphere using two adjectives and suggest a possible event type. The base VLM gives only a brief response (``social and lively'') and does not suggest an event type, whereas the merged VLM produces a more detailed answer with well-chosen adjectives (``elegant'' and ``lively'') and proposes plausible event types such as a corporate party, product launch, or networking event.

\textbf{Math:} As discussed in the main paper and supported by the full MathVista results in Appendix~\ref{sec:mathvista}, merging often yields modest improvements on the reasoning-heavy ``Math'' subset while hurting the perception-heavy ``General'' subset. This suggests that mathematical reasoning does transfer, although some of the gains are offset by degraded visual understanding. In our MathVista evaluation, merged models perform better on reasoning-intensive questions such as ``Find the length of $AC$ in the isosceles triangle $ABC$,'' which the base model fails to answer correctly, yet they can lose the ability to answer visually dependent questions such as ``Does Aqua have the minimum area under the curve?'' The examples in Table~\ref{tab:qualitative_math_qwen72} illustrate this pattern.

\begin{table*}[p]
\centering
\small
\setlength{\tabcolsep}{0pt}
\renewcommand{\arraystretch}{1.1}
\begin{tabularx}{\textwidth}{@{}p{0.40\textwidth}@{\hspace{0.02\textwidth}}X@{}}
\toprule
\multicolumn{2}{@{}l@{}}{\textbf{Example 1: Qwen2-VL-72B in instruction-following}} \\
\midrule
\begin{minipage}[t]{\linewidth}
\vspace{0pt}
\centering
\includegraphics[width=0.90\linewidth]{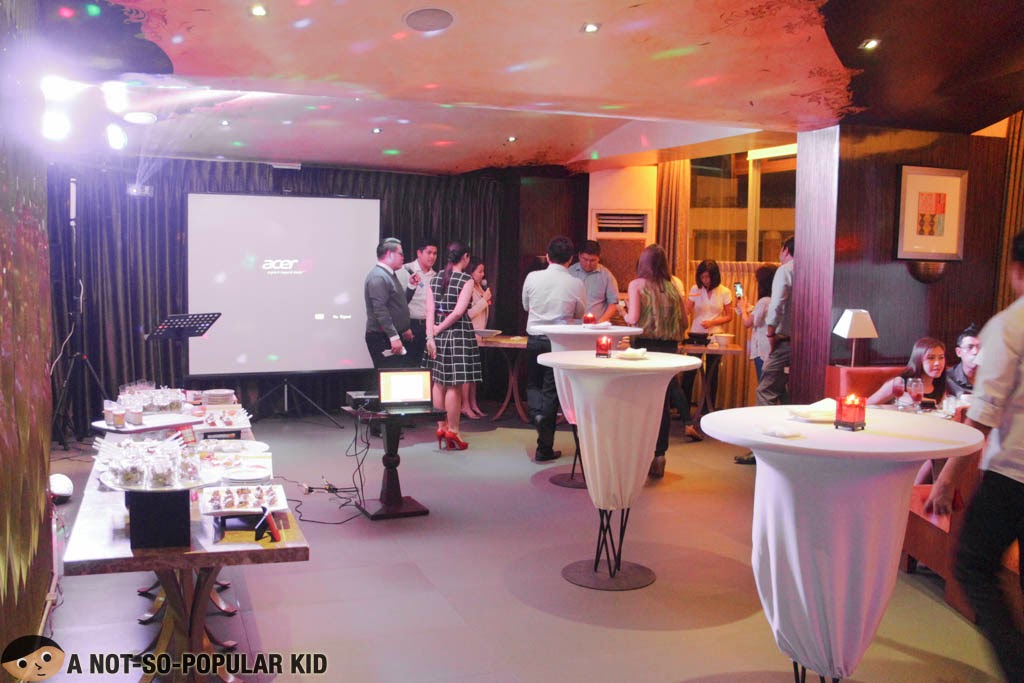} \\[0.45em]
\raggedright
\textbf{Input.} Give an analysis of the atmosphere conveyed in the picture using two adjectives, and suggest what type of event it might be.
\end{minipage}
&
\begin{minipage}[t]{\linewidth}
\vspace{0pt}
\textbf{Base VLM.} ``The atmosphere of the image is social and lively.'' \\
\textit{Score:} Component 1: 4/4; Component 2: 2/3; Component 3: 0/3; Total: 6/10. The response provides two adjectives but does not suggest an event type. \\

\vspace{0.45em}
\textbf{Merged VLM.} ``The atmosphere conveyed in the picture can be described as elegant and lively. The room is decorated with white tablecloths, candles, and a projector screen, giving it a sophisticated feel. At the same time, the colorful lighting and the presence of people mingling suggest a festive atmosphere. This type of event could be a corporate party, a product launch, or a networking event.'' \\
\textit{Score:} Component 1: 4/4; Component 2: 3/3; Component 3: 3/3; Total: 10/10. The response uses exactly two adjectives (``elegant'' and ``lively'') and suggests plausible types of events, satisfying all requirements.
\end{minipage} \\
\bottomrule
\end{tabularx}
\caption{A example from MIA-Bench. The reasoning process of the judge model is summarized. The merged model follows the compositional instruction more completely than the base VLM by both selecting two suitable adjectives and inferring plausible event types from the scene. }
\label{tab:qualitative_party}
\end{table*}

\begin{table*}[p]
\centering
\small
\setlength{\tabcolsep}{6pt}
\renewcommand{\arraystretch}{1.1}
\begin{tabularx}{\textwidth}{@{}>{\raggedright\arraybackslash}X|>{\raggedright\arraybackslash}X@{\hspace{0.012\textwidth}}>{\raggedright\arraybackslash}X@{}}
\toprule
\multicolumn{3}{@{}l@{}}{\textbf{Example 2: Qwen2-VL-72B in mathematical reasoning}} \\
\midrule
\multicolumn{1}{c|}{\textbf{Merged incorrect}} &
\multicolumn{2}{c@{}}{\textbf{Merged correct}} \\
\cmidrule(r{0.8em}){1-1}\cmidrule(l{0.8em}){2-3}
\textbf{Example A: Visual comparison failure} &
\textbf{Example B: Simple function reading} &
\textbf{Example C: Geometric reasoning gain} \\
\begin{minipage}[t]{\linewidth}
\vspace{0pt}
\centering
\begin{minipage}[c][0.17\textheight][c]{\linewidth}
\centering
\includegraphics[width=0.82\linewidth,height=0.14\textheight,keepaspectratio]{aqua.png}
\end{minipage} \\[0.35em]
\raggedright
\textbf{Question.} Does Aqua have the minimum area under the curve? \\
\textit{Gold:} No \\
\textbf{Base VLM.} B. No (\textbf{Correct}) \\
\textbf{Merged VLM.} A. Yes (\textbf{Wrong}) \\
\textit{Interpretation:} This question is perception-heavy: the model must compare multiple plotted curves, and the merged model still fails on this visually judgment.
\end{minipage}
&
\begin{minipage}[t]{\linewidth}
\vspace{0pt}
\centering
\begin{minipage}[c][0.17\textheight][c]{\linewidth}
\centering
\includegraphics[width=0.72\linewidth,height=0.14\textheight,keepaspectratio]{f3.png}
\end{minipage} \\[0.35em]
\raggedright
\textbf{Question.} Is $f(3) > 0$? \\
\textit{Gold:} Yes \\
\textbf{Base VLM.} B. No (\textbf{Wrong}) \\
\textbf{Merged VLM.} A. Yes (\textbf{Correct}) \\
\textit{Interpretation:} Once the graph is localized correctly, the remaining step is a simple sign judgment; the merged model handles this reasoning step better.
\end{minipage}
&
\begin{minipage}[t]{\linewidth}
\vspace{0pt}
\centering
\begin{minipage}[c][0.17\textheight][c]{\linewidth}
\centering
\includegraphics[width=0.86\linewidth,height=0.14\textheight,keepaspectratio]{tra.png}
\end{minipage} \\[0.35em]
\raggedright
\textbf{Question.} Find the length of $AC$ in the isosceles triangle $ABC$. \\
\textit{Gold:} 7 \\
\textbf{Base VLM.} C (\textbf{Wrong}) \\
\textbf{Merged VLM.} B. 7 (\textbf{Correct}) \\
\textit{Interpretation:} The figure provides the setup, but the core difficulty is algebraic reasoning from the isosceles constraint, where merging is beneficial.
\end{minipage} \\
\bottomrule
\end{tabularx}
\caption{Math examples for Qwen2 (72B). The merged model improves on reasoning-centric math questions such as function-value judgment and geometric inference, but it can still fail on perception-heavy visual comparison. This supports our claim that some mathematical reasoning transfers, while visually understanding remains a bottleneck.}
\label{tab:qualitative_math_qwen72}
\end{table*}

\clearpage

\section{Efficiency Comparison of Merging Methods}
\label{sec:efficiency}

Table~\ref{tab:efficiency} reports the per-configuration merge time measured on a single A800 GPU under the Chinese ability injection setup, with Llava as the base VLM and a Mistral model fine-tuned on Chinese data as the expert LLM. For classic methods, we use MergeKit~\cite{mergekit}; for the remaining methods, we use our own implementations. Actual runtime may vary with implementation details and, for data-aware methods, with the size of the calibration dataset.

At the per-configuration level, classic methods and NaN are the most efficient, each requiring only a little over three minutes. MetaGPT also remains in the low-cost regime, although it is slightly slower due to its more complex computation. By contrast, subspace-based tuning-free methods such as WUDI and TSV incur substantially higher overhead, as they require additional subspace construction and transformation. Data-aware methods are more expensive still, because they must compute statistics from calibration data rather than derive merged weights directly from model parameters.

It is important to distinguish per-configuration cost from end-to-end cost, which also includes hyperparameter tuning. As shown in Table~\ref{tab:cost}, classic methods still require substantially more total time to reach their best results because they rely on searching over multiple configurations.

\begin{table}[htbp]
\centering
\small
\begin{tabularx}{\columnwidth}{Xr}
\toprule
\textbf{Method} & \textbf{Merge Time (s)} \\
\midrule
\multicolumn{2}{l}{\textit{Classic}} \\
\quad TA   & 196.529 \\
\quad DARE & 195.360 \\
\quad TIES & 194.995 \\
\midrule
\multicolumn{2}{l}{\textit{Data-aware}} \\
\quad Fisher & 9461.564 \\
\quad RegMean& 3489.240 \\
\midrule
\multicolumn{2}{l}{\textit{Tuning-free}} \\
\quad WUDI   & 1085.351 \\
\quad TSV    & 2676.234 \\
\quad MetaGPT& 259.845 \\
\quad NaN    & 195.667 \\
\bottomrule
\end{tabularx}
\caption{Per-configuration merge time of different methods under the Chinese ability injection setting. Classic methods and NaN are the most efficient, with MetaGPT being slightly slower; WUDI and TSV are intermediate; data-aware methods are the slowest.}
\label{tab:efficiency}
\end{table}

\section{Validation-Test Consistency Analysis}

\label{sec:valtest}

To validate the reliability of using validation sets for hyperparameter optimization, we analyze the correlation between optimization trajectories on the validation and test sets (based on the grid described in Appendix~\ref{sec:hyper}). Figure~\ref{fig:val_test} illustrates the normalized regret curves for DARE across CMMMU dataset.

\begin{figure}[htbp]
    \centering
    \includegraphics[width=\linewidth]{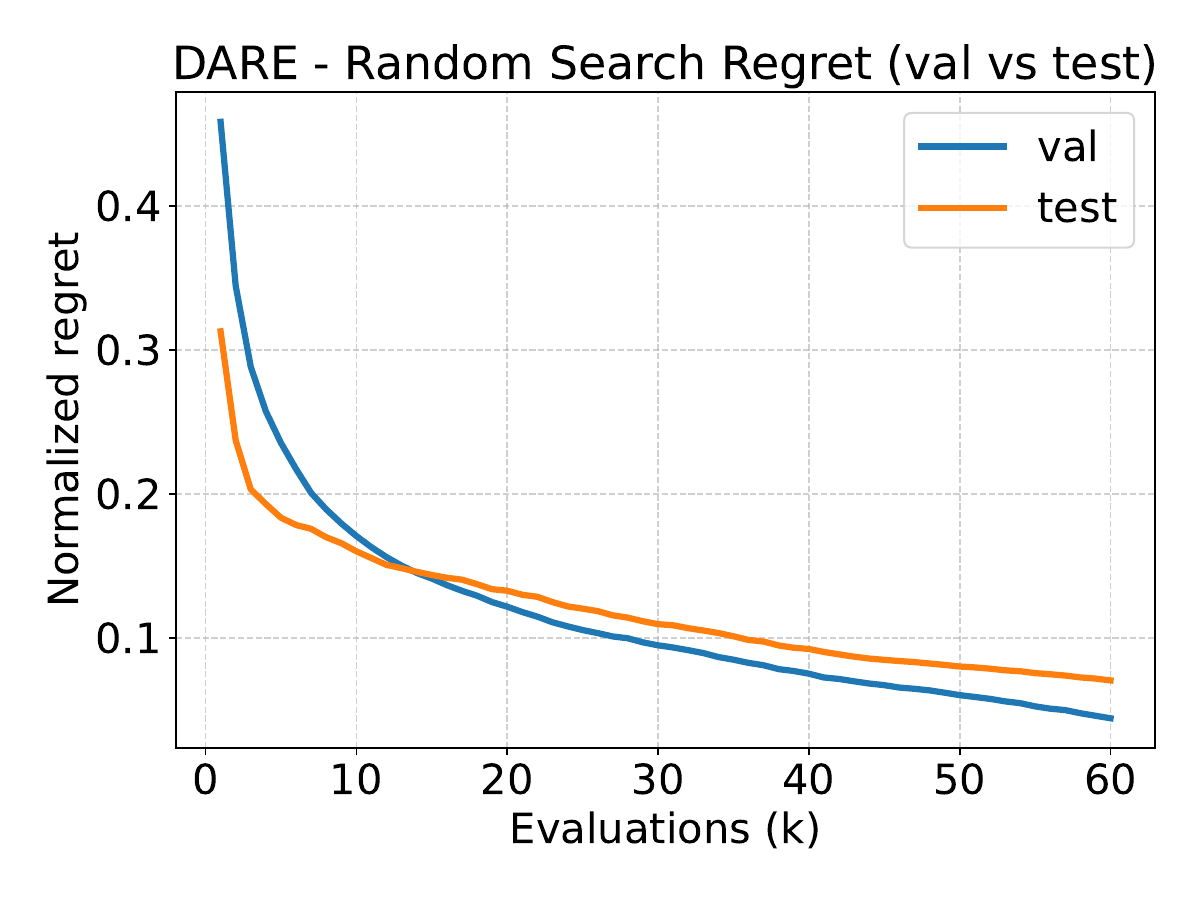}
    \caption{Normalized regret curves for validation and test sets across hyperparameter configurations. The closely aligned trajectories indicate that validation-based optimization effectively generalizes to unseen test data.}
    \label{fig:val_test}
\end{figure}

The optimization trajectories exhibit strong agreement between validation and test sets across all scenarios. As the number of evaluations increases, normalized regret decreases consistently on both sets, with the validation set serving as a faithful proxy for test performance. 

This close correspondence justifies our experimental design: by conducting hyperparameter search exclusively on the 20\% validation split, we avoid the prohibitive expense of repeated test-set evaluations while maintaining the reliability of our conclusions.

\section{Surrogate Model Fidelity Analysis}
\label{sec:surrogate_validity}

Running a comprehensive Hyperparameter Optimization (HPO) benchmark across multiple optimizers, datasets, and merging methods is computationally expensive and would incur substantial API cost if every trial were evaluated directly. We therefore approximate the hyperparameter landscape with a surrogate based on Radial Basis Function (RBF) interpolation, which provides a continuous and query-efficient proxy for the true evaluation process.

To assess fidelity, we perform Leave-One-Out Cross-Validation (LOOCV). For each dataset--method pair, we randomly withhold one interior grid point (excluding boundary points), fit the RBF surrogate on the remaining points, and measure the prediction error on the held-out point. We repeat this process 20 times with different random seeds and report Root Mean Square Error normalized by the metric range (RMSE \%).

Table~\ref{tab:surrogate_rmse} shows that RMSE is below 10\% for most settings and remains mostly within 10--20\% otherwise. This suggests that the surrogate captures the global structure of the merging landscape well. In higher-RMSE cases, it mainly smooths local noise or sharp fluctuations while preserving broad trends such as the location of the main optimum basin. This level of fidelity is sufficient for our goal of comparing optimizers by their ability to find strong regions of the landscape rather than fit every local irregularity.

\begin{table}[htbp]
\centering
\small
\renewcommand{\arraystretch}{1.1}
\setlength{\tabcolsep}{8pt}
\begin{tabularx}{\columnwidth}{l >{\centering\arraybackslash}X >{\centering\arraybackslash}X >{\centering\arraybackslash}X}
\toprule
\textbf{Dataset} & \textbf{DARE} & \textbf{TA} & \textbf{TIES} \\
\midrule
CMMMU     & 9.27  & 7.77  & 15.99 \\
JMMMU     & 6.96  & 18.79 & 8.72 \\
MV-L3     & 17.42 & 14.75 & 9.24 \\
MV-M      & 9.55  & 17.53 & 14.06 \\
MVis-L    & 10.25 & 20.35 & 11.36 \\
MVis-M    & 11.15 & 16.69 & 11.91 \\
MIA-I2    & 4.58  & 5.63  & 7.65 \\
MIA-Q2    & 4.60  & 11.37 & 4.45 \\
WV-I2     & 7.62  & 11.47 & 12.22 \\
WV-Q2     & 8.26  & 13.16 & 4.58 \\
\bottomrule
\end{tabularx}
\caption{Interpolation RMSE (\%) for the surrogate model, averaged over 20 runs using a leave-one-out protocol. DARE refers to DARE-Linear. Lower values indicate higher fidelity. Even higher values ($\sim$15\%) are acceptable as they reflect a smoothing of local noise while preserving global landscape structure. Abbreviations—MV: MathVerse, MVis: MathVista, MIA: MIA-Bench, WV: WildVision; L3: LLaMA3, M: Mistral, L: LLaMA, I2: Idefics2, Q2: Qwen2.}
\label{tab:surrogate_rmse}
\end{table}

\clearpage
\end{document}